\documentclass[journal]{vgtc}                     % final (journal style)

\onlineid{8402}

\vgtccategory{Research\vspace{-10pt}}

\title{A Perturbation-Driven Framework for Sensitivity Analysis of Dimensionality Reduction\vspace{-4pt}}

\author{%
  \authororcid{Md Afridi Hasan}{0009-0000-9581-408X},
  \authororcid{Jaemin Jo}{0000-0002-5207-6010}, and
  \authororcid{Takanori Fujiwara}{0000-0002-6382-2752}
}

\authorfooter{
  
  \item
  	Md Afridi Hasan and Takanori Fujiwara are with University of Arizona.
  	E-mail: \{afridi, tfujiwara\}@arizona.edu
  \item
  	Jaemin Jo is with Sungkyunkwan University.
  	E-mail: jmjo@g.skku.edu.
}

\abstract{%
  Low-dimensional representations produced by dimensionality reduction (DR) are commonly evaluated using distortion measures such as trustworthiness and continuity. In this paper, we study the sensitivity of DR methods to plausible input variations by quantifying changes in structural preservation and visually diagnosing how these changes manifest in the resulting representations. This perspective reflects the fact that real-world datasets are finite, noisy samples of an underlying distribution, and DR should not produce substantially different structures under small data changes. We introduce a model-agnostic, perturbation-driven sensitivity evaluation framework that treats perturbations as controlled probes of plausible data variation. Across multiple perturbation families and strength levels, we repeatedly perturb the input, apply a DR method, and evaluate the resulting low-dimensional representations using a set of distortion measures to derive sensitivity profiles for each DR method. We demonstrate the utility of our framework through case studies on DR method comparison, behavior analysis, and robustness improvement. We find that methods performing well according to conventional distortion measures (e.g., t-SNE) can still exhibit substantial sensitivity, leading to instability in visually salient structures such as trajectories and clusters, while certain methods (e.g., multidimensional scaling) may exhibit improved structural preservation under specific perturbations. These results highlight the importance of sensitivity analysis for comparing DR methods, diagnosing perturbation-sensitive visual patterns, and guiding the development of more robust DR methods toward trustworthy visual analytics.
  \vspace{-2pt}
}


\graphicspath{{figs/}{figures/}{pictures/}{images/}{./}} % where to search for the images

\usepackage{tabu}                      % only used for the table example
\usepackage{booktabs}                  % only used for the table example
\usepackage{lipsum}                    % used to generate placeholder text
\usepackage{mwe}                       % used to generate placeholder figures
\usepackage{ccicons}                   % package to be able to use icons from creative commons
\usepackage{xcolor} 
\usepackage[normalem]{ulem}

\usepackage{mathptmx}                  % use matching math font
\usepackage{amsfonts}

\usepackage{rsfso}

\usepackage{comment}

\usepackage{tikz}

\usepackage{multicol}
\renewcommand{\manuscriptnotetxt}{}

\usepackage[font={scriptsize,sf}]{caption}
\usepackage{setspace}

\begin{document}

%%%%%%%%%%%%%%%%%%%%%%%%%%%%%%%%%%%%%%%%%%%%%%%%%%%%%%%%%%%%%%%%
%%%%%%%%%%%%%%%%%%%%%% START OF THE PAPER %%%%%%%%%%%%%%%%%%%%%%
%%%%%%%%%%%%%%%%%%%%%%%%%%%%%%%%%%%%%%%%%%%%%%%%%%%%%%%%%%%%%%%%

\firstsection{Introduction}

\maketitle

\urlstyle{rm}
\newcommand{\OrgMat}{\mathbf{X}}
\newcommand{\DRResult}{\mathbf{Z}}
\newcommand{\nInsts}{N}
\newcommand{\nAttrs}{D}
\newcommand{\nLatFeats}{{D'}}
\newcommand{\nRuns}{R}
\newcommand{\DistortionScore}{s}
\newcommand{\DistortionScores}{\mathbf{s}}
\newcommand{\Labels}{\mathbf{y}}
\newcommand{\MeasureSummary}{\mathcal{S}}
\newcommand{\PerturbStrength}{\epsilon}
\newcommand{\nPerturbStrengths}{T}
\newcommand{\VisualSummary}{\mathcal{V}}

High-dimensional data are ubiquitous in visual analytics, arising in domains such as bioinformatics, natural language processing, and computer vision~\cite{liu2016visualizing,xuan2025attributionscanner,huertas2023exploring,kobak2021initialization}. 
Analysts frequently use dimensionality reduction (DR) to construct 2D or 3D visual representations (also referred to as projections or embeddings) that support the identification of clusters, trends, trajectories, and outliers~\cite{hyeon2025unveiling,nonato2018multidimensional,van2008visualizing,mcinnes2018umap}.
In these representations, the spatial proximity of instances is commonly interpreted as a proxy for their similarity, enabling analysts to form and test hypotheses about structure in the underlying dataset~\cite{kruskal1964multidimensional,venna2006local,cashman2025critical}.

However, when a dataset's intrinsic dimensionality exceeds the target dimensionality, not all relationships can be preserved exactly.
The resulting distortions may obscure meaningful patterns or suggest structures that are not supported by the data~\cite{wattenberg2016use,machado2025necessary}. 
Because these representations are often treated as visual evidence in exploratory analysis, assessing their quality and reliability is critical for establishing calibrated trust in visual interpretation ~\cite{cashman2025critical,sacha2015role,hullman2018pursuit}.

A substantial body of work has therefore developed \textit{distortion measures} (also referred to as projection quality metrics) to evaluate low-dimensional representations~\cite{bertini2011quality,venna2006local,lee2009quality}. 
Each measure compares a dataset with its representation and returns a distortion-measure value quantifying the preservation of a particular structural property, such as global relationships, local neighborhoods, or cluster-level structure~\cite{kruskal1964multidimensional,lee2009quality,sips2009selecting,jeon2024classes}. 
Multiple measures are often used together to capture complementary aspects of structural preservation~\cite{bertini2011quality,espadoto2021toward,machado2025necessary}. 
However, conventional evaluation applies these measures to a fixed dataset--representation pair or summarizes their values across repeated runs on the same dataset.

Many real-world datasets can be viewed as finite, potentially noisy samples of an underlying distribution. 
Small variations in the observed data may alter the resulting DR representations~\cite{garcia2013stability,johnson2022embedr,zabel2025visualizing,sun2023dynamic}, and thereby affect conclusions drawn from them~\cite{potter2011quantification,pang1997approaches,hullman2018pursuit}.
We therefore ask: \textit{How consistently does a DR method preserve relevant structural relationships when the input data changes slightly?}
We investigate this by tracking distortion-measure responses to controlled input perturbations.

The sensitivity of DR representations to small input variations is recognized in practice through visual examination under controlled perturbations~\cite{bredius2022visual,faust2018dimreader} and is also reflected in dynamic DR methods and measures for cross-layout stability~\cite{wang2022joint,vernier2020quantitative}.
Prior work has analytically characterized sensitivity for particular DR methods~\cite{liu2025assessing,zabel2025visualizing} and quantitatively evaluated layout stability under repeated runs, bootstrap resampling, or controlled changes to input data and analysis configurations~\cite{hamad2018stability,sun2023dynamic,atzberger2025large}. 
These studies show that DR results can be sensitive to input changes, but empirical evaluations typically focus on a particular perturbation mechanism or application domain.
Sample-composition variation, outlier influence, and observational noise represent distinct, plausible forms of data variation that can expose different weaknesses in a DR method. Robustness under one therefore does not imply robustness under the others.
Accordingly, prior work does not provide a common, model-agnostic basis for evaluating DR methods under multiple plausible forms of data variation and with respect to complementary structural properties.
Such a basis would support consistent method comparison and selection, diagnosis of affected structures, and robustness improvements targeting perturbation-specific weaknesses.

To address this gap, we introduce a \textit{model-agnostic, perturbation-driven sensitivity evaluation framework} that connects sensitivity analysis to established distortion measures~\cite{machado2025necessary,jeon2023zadu}. 
Treating \textit{perturbations as controlled probes} of plausible data variation, we generate datasets under diverse perturbation strategies and strengths, apply DR methods to each dataset, and compute the resulting distortion-measure values. We summarize and visualize these values as \textit{sensitivity profiles}, which show how distortion-measure values vary with perturbation type and strength. 
We also provide instance-level diagnostic visualizations to further reveal where sensitivity occurs and relate variation in per-instance measure values to visually salient structures in the representations.

Our empirical evaluation through case studies demonstrates the value of perturbation-driven assessment. 
First, sensitivity profiles reveal changes in structural preservation that baseline evaluation alone cannot show, supporting context-specific DR method selection. For example, t-SNE performs well on the unperturbed data but can still be sensitive to particular perturbations.
Second, instance-level analysis and inspection of the resulting representations localize sensitivity to visually salient structures, including trajectories, clusters, and regions influenced by outliers, helping analysts identify interpretations that warrant qualification or further investigation.
Third, sensitivity analysis can guide the evaluation and design of robust DR methods, as illustrated by comparison with robust principal component analysis (PCA) and our proof-of-concept multidimensional scaling (MDS) variant.

In summary, our primary contributions are:
\begin{itemize}[topsep=0pt,parsep=0pt,leftmargin=17pt]
  \item A model-agnostic framework for characterizing changes in structural preservation across controlled perturbation methods.
  \item Distortion-measure-based sensitivity profiles and diagnostic visualizations that characterize changes in local, global, and cluster-level preservation and relate them to structures in DR representations. 
  \item  Case studies demonstrating how sensitivity analysis can support DR method comparison, diagnosis of perturbation-specific behavior, and evaluation and design of robustness-oriented DR variants.
\end{itemize}

\section{Background and Related Work}
\vspace{-1pt}
We review DR, distortion measures, input-perturbation sensitivity analysis, and temporal stability measures relevant to our framework

\vspace{-1pt}
\subsection{Dimensionality Reduction (DR)}
\vspace{-1pt}

DR transforms a high-dimensional dataset, $\smash{\OrgMat{} \in \mathbb{R}^{\nInsts \times \nAttrs}}$, into its low-dimensional representation, $\smash{\DRResult{} \in \mathbb{R}^{\nInsts \times \nLatFeats}}$, where $\nInsts$, $\nAttrs$, and $\nLatFeats$ denote the numbers of instances, attributes, and latent features, respectively. 

Each DR method has a specific optimization objective to preserve essential properties of the original data. 
For example, PCA~\cite{hotelling1933analysis,jolliffe1986principal} maximizes the variance of transformed data; MDS~\cite{torgerson1952} minimizes the discrepancy between pairwise distances in $\OrgMat{}$ and $\DRResult{}$; and neighbor embedding methods such as t-SNE~\cite{van2008visualizing} and UMAP~\cite{mcinnes2018umap} emphasize preservation of local neighborhood structure. 
Surveys provide broader comparisons of DR methods~\cite{van2009dimensionality,cunningham2015linear,espadoto2021toward}.

When relevant high-dimensional relationships cannot be represented exactly in the target dimensionality, DR may obscure meaningful patterns or introduce spurious structures~\cite{hyeon2025unveiling,liu2025assessing,cashman2025critical}.

\vspace{-1pt}
\subsection{Distortion Measures}
\vspace{-1pt}

Distortion measures~\cite{espadoto2021toward,jeon2023zadu,machado2025necessary} have been developed to evaluate DR methods, guide the selection of a method suitable for a given task (e.g., cluster identification), and assess the reliability of low-dimensional representations. 
Most distortion measures are model-agnostic and compare $\OrgMat{}$ with $\DRResult{}$ to quantify a particular aspect of structural preservation.

According to their structural focus~\cite{jeon2021measuring}, distortion measures can assess global, local, or cluster-level relationships. 
Global-level measures consider all pairwise relationships among instances. 
For example, \texttt{normalized stress} quantifies the discrepancy between pairwise distances in $\OrgMat{}$ and $\DRResult{}$. 
Local-level measures evaluate the preservation of neighborhood structure. 
For instance, \texttt{trustworthiness}~\cite{venna2001neighborhood} measures the extent to which the $k$-nearest neighbors of each instance in $\DRResult{}$ are also neighbors in $\OrgMat{}$. 
Conversely, \texttt{continuity} measures the extent to which the $k$-nearest neighbors in $\OrgMat{}$ remain neighbors in $\DRResult{}$. 
Cluster-level measures extend local-level measures to assess cluster structure preservation. 
For example, \texttt{steadiness}~\cite{jeon2021measuring} extends \texttt{trustworthiness} and indicates how well clusters in $\DRResult{}$ are preserved in $\OrgMat{}$, while \texttt{cohesiveness} extends \texttt{continuity}.
Because distortion measures capture different structural properties, multiple measures are often used together to characterize complementary strengths and weaknesses of DR methods~\cite{espadoto2021toward}.
Conventionally, distortion measures compare a fixed dataset--representation pair and do not characterize how distortion-measure values change when the input dataset is slightly perturbed.
Consequently, evaluating only an unperturbed dataset may mask a DR method's sensitivity to variability introduced during data collection or sampling~\cite{sun2023dynamic}.
We quantify this sensitivity by tracking distortion-measure values across controlled families of related datasets.

\vspace{-1pt}
\subsection{Sensitivity Analysis of DR Methods}
\vspace{-1pt}

Researchers have investigated the sensitivity of DR methods,  
including sensitivity to initialization of the low-dimensional representation prior to optimization~\cite{kobak2021initialization}, stochastic optimization processes~\cite{jung2024ghostumap,jung2025ghostumap2}, and data scaling~\cite{lehmann2015optimal,fujiwara2023feature,green1975robustness}.
Here, we focus on discussing the sensitivity of DR methods to small perturbations of input datasets.

Prior analyses have examined PCA's sensitivity to outliers~\cite{serneels2008principal}, MDS's response to perturbations in distance relationships~\cite{sibson1979studies}, and t-SNE's sensitivity to small input changes~\cite{zabel2025visualizing}. 
Some of these analyses derive sensitivity from the mathematical formulation of a particular DR method, for example, through Jacobians with respect to the high-dimensional input.
Although precise, such analyses are model-specific and therefore difficult to apply uniformly across DR methods.

Several empirical studies are closely related to ours in examining how DR representations respond to controlled changes. 
DynamicViz~\cite{sun2023dynamic} uses bootstrap resampling to simulate sampling variability, repeatedly applies a given DR method, and quantifies the variance of normalized interpoint distances across the resulting representations. 
Related reliability methods, including EMBEDR~\cite{johnson2022embedr} and scDEED~\cite{xia2024statistical}, also use input perturbations to identify unreliable relationships in low-dimensional representations.
Bredius et al.~\cite{bredius2022visual} applied feature perturbations at varying strengths to a pretrained neural-network DR model and visually identified the point when its cluster structure ceased to resemble the original representation.
Atzberger et al.~\cite{atzberger2025large} quantitatively analyzed text-spatialization stability under input jitter strengths, hyperparameter changes, and random initialization using local, global, and class-separation similarities between pairs of layouts.

These studies demonstrate the value of examining DR results under controlled changes, but individually focus on particular perturbation mechanisms, application domains, or DR methods.
Our framework instead evaluates each perturbed dataset against its corresponding representation using established distortion measures and tracks changes in these measures across multiple perturbation types and strengths. 
This operationalization supports model-agnostic sensitivity analysis even when perturbations alter sample composition.

\subsection{Temporal Stability Measures for Dynamic DR}
\label{sec:dynamic_dr}

Dynamic DR methods generate comparable layouts for evolving high-dimensional datasets~\cite{fujiwara2019incremental,rauber2016visualizing}.
We focus on their cross-layout stability measures rather than comprehensively reviewing these DR methods.

Dynamic t-SNE~\cite{rauber2016visualizing} introduces a temporal penalty that discourages large movements of corresponding instances between consecutive layouts, although its stability was assessed through visual inspection.
Joint t-SNE~\cite{wang2022joint} produces comparable projections by preserving neighborhoods that remain similar across datasets.
It also uses local coherence error (LCE) to quantify changes in the relative positions of instances within these stable neighborhoods. 
Vernier et al.~\cite{vernier2020quantitative} introduced more general measures, including temporal stress and temporal Shepard correlations, that compare how much each instance changes in the high-dimensional space with how far it moves in the projection.

These temporal measures can complement our framework when one-to-one instance correspondence is preserved, as under attribute-noise perturbations (cf.~\cref{sec:noise_addition}). 
Our framework also considers sampling and outlier perturbations that remove or add instances. 
Since these perturbations do not preserve full correspondence, the temporal measures are not directly applicable.
Moreover, whereas these temporal measures primarily characterize positional or relational changes across layouts, our framework uses distortion measures to evaluate structural preservation within each dataset–representation pair and tracks how the resulting values change with perturbation strength.

\vspace{3pt}
\noindent 
\textbf{Positioning of our work.}
Overall, prior studies have examined DR stability using perturbations or related datasets, but typically focus on particular perturbation mechanisms, application domains, DR methods, or direct comparisons between layouts. 
Our framework combines multiple perturbation methods and strength sweeps with established local, global, and cluster-level distortion measures.
It thereby produces sensitivity profiles that reveal how sensitivity depends on the DR method, dataset, and perturbation configuration while enabling diagnosis of the structures associated with observed changes.

\begin{figure*}[t]
    \centering
    \includegraphics[
        alt={},
        width=\linewidth
    ]{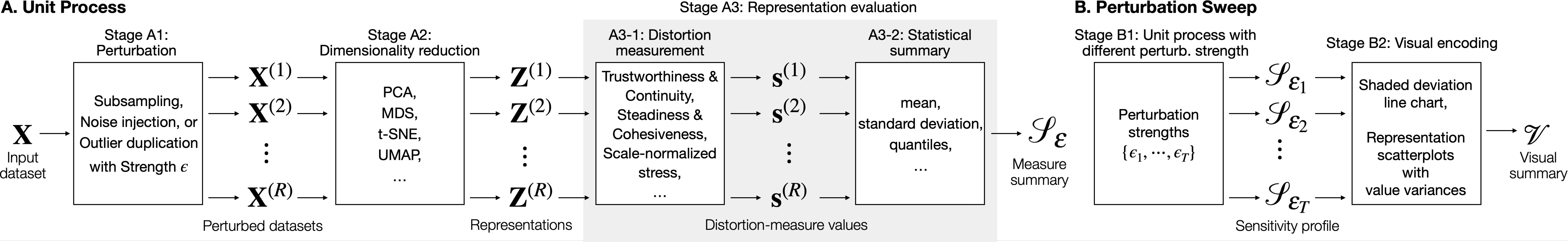}
    \caption{A perturbation-driven sensitivity evaluation pipeline: (A) the unit process computes and summarizes distortion-measure values under a fixed perturbation condition, and (B) the perturbation sweep repeats this process across perturbation strengths to construct sensitivity profiles.
    }
    \label{fig:pipeline}
\end{figure*}

\section{Usage} Scenarios
\label{sec:motivating_scenarios}

We describe three scenarios illustrating how our framework can support visual analytics with DR. 
They concern evaluating whether structural preservation is sensitive to plausible variations in an observed dataset, including sampling variability, observational noise, and outlier contamination.
Such variations are not examined by conventional evaluations that treat datasets as fixed. 
These scenarios differ from dynamic DR settings, where changes represent meaningful temporal evolution and cross-layout stability is a separate evaluation objective~\cite{rauber2016visualizing,vernier2020quantitative,wang2022joint}. 
Case studies corresponding to the scenarios are presented in \cref{sec:case_studies}.

\vspace{3pt}\noindent\textbf{Scenario 1: Selecting a DR method for a given dataset and analysis task.}
An analyst needs to choose a DR method for a downstream task such as cluster identification or anomaly detection~\cite{cashman2025critical,hyeon2025unveiling}.
The analyst first identifies plausible sources of data variation, such as sampling variability, observational noise, or outlier contamination, and selects distortion measures aligned with the task. 
Values computed on the unperturbed dataset provide a baseline, while sensitivity profiles show how these values respond to the selected perturbations.
For instance, one method may achieve higher \texttt{trustworthiness} on the unperturbed dataset but degrade substantially under sampling variation, while another may have slightly lower initial \texttt{trustworthiness} but show less degradation as perturbations intensify. 
The profiles support method selection based on baseline performance and sensitivity to perturbations relevant to the analysis.
The selection should consider both structural preservation and its sensitivity to application-relevant input variations. Better structural preservation under these variations can favor a method, but substantial sensitivity suggests a need to examine whether the structures supporting the intended analysis remain consistent. Sensitivity profiles and instance-level diagnostic visualizations support this assessment by showing changes in structural preservation and helping identify affected regions or structures.

\vspace{3pt}\noindent\textbf{Scenario 2:
Diagnosing DR behavior and qualifying an insight from DR visualizations.}
An analyst derives insights from a DR visualization, such as identifying clusters or relationships between groups~\cite{packer2019lineage,fujiwara2019supporting,hyeon2025unveiling}. 
The analyst selects distortion measures corresponding to the structural relationships underlying the interpretation and examines how their values change under plausible perturbations. 
Per-instance values and the resulting representations help localize this variability.

If the relevant measures remain high and consistent and visual inspection shows that the pattern of interest persists, the analysis provides additional support for the interpretation. Conversely, degradation or high variability in the measures, together with visible changes in the corresponding structure, indicates that the interpretation warrants further examination. 
The analyst may then inspect the original high-dimensional relationships, compare alternative methods or configurations, or qualify the reported finding as sensitive to data variation. 
Thus, the framework provides diagnostic evidence rather than independently establishing that a visual pattern is correct, readable, or meaningful.

\vspace{3pt}\noindent\textbf{Scenario 3: Guiding robustness improvement of DR methods.} 
A researcher developing or adapting a DR method seeks to make its preservation of relevant structural relationships less sensitive to input variations relevant to the intended application~\cite{xu2010robust,candes2011robust,green1975robustness,sibson1979studies}.
Existing evaluations offer limited insight into its response to perturbations.

The researcher can use sensitivity profiles to identify specific weaknesses. 
For example, a method may preserve relevant structures consistently under observational noise but exhibit degradation under outlier inclusion or sampling variation.
These findings can guide modifications to the objective function, neighborhood definition, regularization, or preprocessing. The revised method can then be evaluated under the same conditions to determine whether the targeted sensitivity has been reduced without degrading its performance on the unperturbed data.

\begin{figure*}[t]
    \centering
    \includegraphics[
        alt={},
        width=0.95\linewidth
    ]{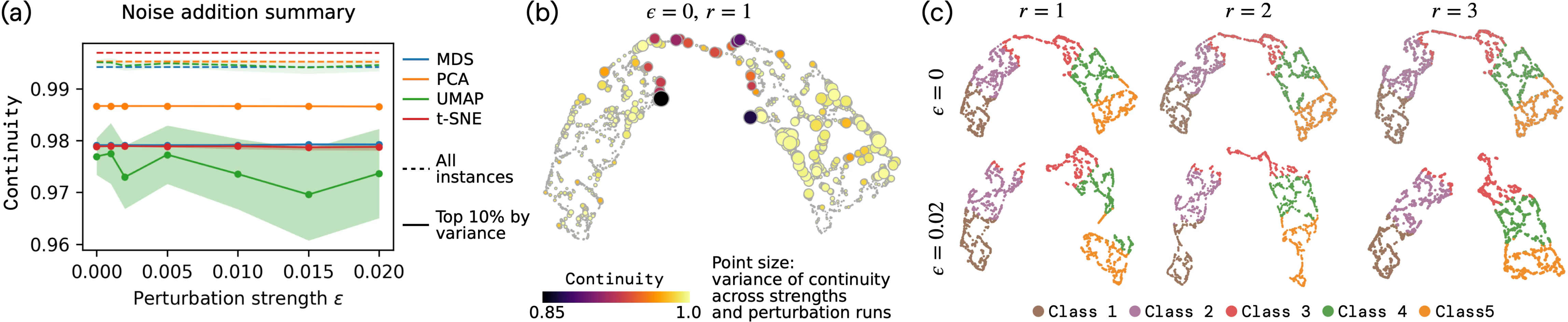}
    \caption{Example usage of our framework on the \texttt{Swiss\,roll} dataset~\cite{tenenbaum2000global}.
    (a) Sensitivity profiles under noise perturbation for each DR method. The dashed line shows the mean \texttt{continuity} across all instances, the solid line shows the mean \texttt{continuity} of the 10\% of instances with the highest variance, and the shaded region indicates the standard deviation across perturbation runs. 
    (b) A UMAP embedding at $\PerturbStrength=0$ and $r=1$, where color encodes per-instance \texttt{continuity} and point size encodes the variance of \texttt{continuity} across perturbation strengths and runs. 
    (c) UMAP embeddings under different perturbation strengths and runs; colors indicate five continuous regions of the \texttt{Swiss\,roll}.}
    \label{fig:swissroll_example}
\end{figure*}

\section{Methodology}
We describe our perturbation-driven sensitivity evaluation framework,\footnote{The source code will be made publicly available upon publication.} introducing its design goals, pipeline, and individual stages.

\subsection{Goals and Design Motivations} 

Our goal is to characterize DR \emph{sensitivity} to \emph{plausible input perturbations} by examining how the preservation of relevant structural relationships changes with perturbation strength and how these changes manifest in the resulting representations.

Three design requirements guide our framework.
First, the evaluation should be \emph{model-agnostic} and applicable across DR methods with different objectives. 
This motivates a standardized perturbation-based approach rather than method-specific analyses derived from individual objective functions.
Second, sensitivity should be characterized in terms of the structural relationships preserved by each representation, rather than only positional differences between layouts. 
We therefore use established \emph{distortion measures}, which can compare each perturbed dataset with its corresponding representation, remain applicable when sample composition changes, and 
separately characterize the preservation of local, global, and cluster-level structure.
This also connects sensitivity analysis with existing DR evaluation practices.
Third, the evaluation should support \emph{diagnosis} in addition to summary comparison by localizing sensitivity to particular instances, regions, or structures in the resulting representations.
This reveals which aspects of structural preservation are sensitive and where sensitivity occurs.

\subsection{Pipeline Overview}
\cref{fig:pipeline} summarizes the pipeline’s two components: (A) the unit process, which computes and summarizes distortion-measure values for a single perturbation condition, and (B) the perturbation sweep, which repeats this process across strengths to construct sensitivity profiles and their visual summaries. 
The framework is independent of the DR method, perturbation method, and distortion measure; these components can be modified without changing the overall evaluation logic.

Two main challenges guide the design of the evaluation pipeline.
First, the framework should incorporate plausible data perturbations (Stage~A1 in \cref{fig:pipeline}) that mimic datasets drawn from the same underlying distribution as the input data, while introducing controlled, subtle variations (cf.~\cref{sec:perturbation}).
Second, 
the framework should characterize structural preservation across multiple representations obtained from perturbed datasets (Stage~A3). 
This evaluation is not straightforward, as DR implementations may involve randomness arising from initialization or stochastic optimization procedures~\cite{mcinnes2018umap,van2008visualizing,jung2024ghostumap}, and it is difficult to directly compare relationships among instances across low-dimensional representations~\cite{fujiwara2023feature}. 
Instead of directly comparing multiple low-dimensional representations, we use established distortion measures to evaluate each perturbed dataset–representation pair (cf.~\cref{sec:measure_changes}). 
This approach also relies on established evaluation measures familiar to DR users.
We describe each component below.

\vspace{3pt}\noindent\textbf{Unit process.}
Starting from an input dataset $\OrgMat$ (and optional labels $\Labels \in \mathbb{Z}^{\nInsts}$), we first apply a perturbation method (e.g., subsampling) with perturbation strength $\PerturbStrength$ to generate a set of perturbed datasets $\{\OrgMat^{(1)}, \dots, \OrgMat^{(\nRuns)}\}$ (Stage~A1), where $\nRuns$ denotes the number of perturbation runs. 
We run the same perturbation method $\nRuns$ times to capture variability in sampled instances. 
Each perturbed dataset is then embedded into a low-dimensional representation $\{\DRResult^{(1)}, \dots, \DRResult^{(\nRuns)}\}$ by a DR method, such as PCA, MDS, t-SNE, or UMAP (Stage~A2).
For DR methods affected by random initialization or stochastic optimization (e.g., UMAP), this stage can be further repeated over multiple random seeds to capture algorithmic variability. 
Next, each representation is evaluated with a panel of distortion measures, producing vectors of distortion-measure values $\{\DistortionScores^{(1)}, \dots, \DistortionScores^{(\nRuns)}\}$ (Stage~A3-1). 
These values are summarized statistically, for example by their mean, standard deviation, and quantiles, to produce a measure summary $\MeasureSummary_\PerturbStrength$ for the given perturbation condition (Stage~A3-2). 
Variation in distortion-measure values across repeated runs can arise from input perturbations, DR randomness, and their interaction. 
Sec.~D of the supplementary materials examines these contributions through a controlled experiment using multiple DR seeds for each perturbed dataset.

\vspace{3pt}\noindent\textbf{Perturbation sweep.}
Repeating the unit process across \(\{\epsilon_1,\ldots,\epsilon_T\}\) produces the sequence \(\{\MeasureSummary_{\epsilon_1}, \ldots, \MeasureSummary_{\epsilon_T}\}\).
From this sequence, we construct a \textit{sensitivity profile} for each distortion measure, showing how its values vary across perturbation strengths (Stage~B1).
These profiles are then visualized, for example, as shaded line charts (e.g., \cref{fig:swissroll_example}(a)), to form a visual summary $\VisualSummary$ that reveals how sensitivity evolves as perturbation strength increases (Stage~B2). 
In this way, sensitivity is represented not as a single value, but as a function of perturbation type and strength. 
As an auxiliary visualization, we provide a scatterplot of the low-dimensional representation with the variance of per-instance measure values (e.g., \cref{fig:swissroll_example}(b)).
This plot indicates which parts of the representation tend to be non-robust across different perturbation strengths, potentially showing misleading patterns.

\vspace{3pt}\noindent\textbf{Running Example.}
Before detailing each component, we illustrate how our framework supports quantitative comparison and visual diagnosis, using the \texttt{Swiss\,roll} dataset~\cite{pedregosa2011scikit} in \cref{fig:swissroll_example}.

We generate a \texttt{Swiss\,roll} dataset with $N=2000$ and embed its original 3D coordinates into a 50D space using a random linear projection, thereby creating a high-dimensional dataset while preserving the underlying manifold structure.
We compare PCA, MDS, t-SNE, and UMAP under noise-addition perturbations.
We examine noise strengths from $\epsilon=0$ (no added noise) to $\epsilon=0.02$. At each strength, we repeat the perturbation and DR application $R=10$ times. 
The formal definition of the noise-addition perturbation is provided in \cref{sec:noise_addition}.
We evaluate each resulting representation using \texttt{continuity}, which measures whether instances that are close in the corresponding high-dimensional dataset remain close in the low-dimensional representation, with 1.0 indicating perfect preservation.
\texttt{Continuity} also provides a value for each instance, referred to as per-instance \texttt{continuity}.

\cref{fig:swissroll_example}(a) presents the resulting sensitivity profiles.
For each DR method, the dashed line shows the mean per-instance continuity across all instances at each perturbation strength.
The solid line shows the mean for the 10\% of instances with the highest \texttt{continuity} variance across perturbation strengths and runs, thereby highlighting instances whose neighborhood preservation is particularly sensitive.
The shaded region indicates the standard deviation across runs.
The profiles show that UMAP exhibits larger changes and run-to-run variation among high-variance instances than the other methods.

To diagnose where this variation occurs, \cref{fig:swissroll_example}(b) maps per-instance sensitivity onto a UMAP representation obtained at $\epsilon=0$.
Point color encodes per-instance \texttt{continuity}, while point size encodes its variance across perturbation strengths and runs.
The visualization highlights sensitive regions near the top-center and bottom-right.

\cref{fig:swissroll_example}(c) further shows UMAP representations generated at $\epsilon=0$ and $\epsilon=0.02$ across three runs.
Colors identify five continuous regions of the \texttt{Swiss\,roll}.
Visual comparison reveals structural changes under noise perturbation, including the breakdown of the bridge-like structure in the red region and increased variation in the green and orange regions.
Together with the sensitivity profiles, these representations indicate that, under this setting, UMAP is more sensitive to noise than the other evaluated methods, particularly in preserving continuous structures.

The following sections detail these perturbation, measurement, and visualization components.

\subsection{Perturbation Design}
\label{sec:perturbation}
 
We design three representative perturbation methods: (1) subsampling/resampling, (2) outlier duplication, and (3) noise addition, each capturing a different source of data variability.
Subsampling/resampling simulates sampling variability inherent in finite samples when drawing instances from an underlying distribution. 
Outlier duplication simulates increased outlier influence by duplicating rare or extreme instances, reflecting scenarios where outliers are overrepresented due to sampling variability. 
Noise addition simulates off-manifold contamination, reflecting measurement noise or data corruption during data acquisition.

Ideally, the underlying data distribution would be available; however, this is rarely the case in practice. 
Thus, we assume that the observed dataset provides a reasonable approximation of the underlying distribution and use it to generate perturbed sample sets.
We further design each method to include a strength parameter that controls the magnitude of change, enabling evaluation of how perturbation effects emerge as this magnitude increases.
Across all perturbation methods, the strength parameter is defined so that larger values correspond to stronger deviations from the original dataset.

While these perturbations cover common sources of variability encountered during data collection~\cite{wasserman2004all,fuller2009measurement,rousseeuw2011robust,kamal2021recent}, we do not claim that they comprehensively capture all real-world scenarios. 
Instead, we design our framework to be flexible and extensible, allowing new perturbation methods to be incorporated based on specific analysis needs.

\subsubsection{Subsampling/Resampling (Sampling Variation)}
To model changes in which observations are available, such as incomplete data or sampling variability, we generate a perturbed dataset $\OrgMat^{(\MakeLowercase{\nRuns})}$, where $r$ denotes the perturbation index, by sampling instances from $\OrgMat$.
For this sampling, the perturbation strength $\PerturbStrength \in [0, 1)$ controls the fraction of instances sampled from the original dataset of size $\nInsts$. 
Specifically, we sample $\lfloor (1 - \PerturbStrength) \nInsts \rfloor$ instances from $\OrgMat$.
Following DynamicViz~\cite{sun2023dynamic}, we allow the same instance to be selected multiple times from $\OrgMat$ (i.e., with replacement), as we consider $\OrgMat$ to represent the underlying distribution. 
Thus, even when $\PerturbStrength = 0$ (i.e., pure resampling), we can observe variation in sampled instances across runs, which corresponds to bootstrap sampling as employed in DynamicViz.
Unlike DynamicViz, our framework supports a range of perturbation strengths.
$\nRuns$ independent replicates are generated at each strength level so that sensitivity reflects the randomness of which points are retained.
Allowing $\PerturbStrength < 0$ extends this design to oversampling. 
For example, $\PerturbStrength = -1$ yields $2 \nInsts$ instances.

\subsubsection{Outlier Duplication (Outlier Influence)}
To examine how strongly a representation is influenced by a small number of atypical observations, we stochastically duplicate selected outlying points and append the duplicates to $\OrgMat$. 
To this end, we first estimate the outlyingness of each instance in $\OrgMat$ using the isolation forest algorithm~\cite{liu2008isolation}, a widely used and model-agnostic method for detecting anomalies in high-dimensional data. 
Other outlier detection methods can be incorporated within our framework.
The isolation forest produces an anomaly score for each instance.
We use these scores as sampling weights, so that instances with higher anomaly scores are more likely to be selected for duplication.
For this sampling, the perturbation strength $\PerturbStrength \geq 0$ controls the fraction of instances duplicated relative to the original dataset of size $\nInsts$. 
Specifically, we duplicate $\lfloor \PerturbStrength \nInsts \rfloor$ instances from $\OrgMat$. 
Consequently, $\OrgMat^{(\MakeLowercase{\nRuns})}$ contains $\nInsts + \lfloor \PerturbStrength \nInsts \rfloor$ instances. 
This perturbation is intended to reveal whether a method is disproportionately affected by a few high-influence observations.

\vspace{-2pt}
\subsubsection{Noise Addition (Off-manifold Contamination)}
\label{sec:noise_addition}
\vspace{-1pt}

Unlike the above perturbations, noise addition perturbs feature values rather than modifying the set of instances.
To model observational noise and off-manifold contamination, we augment $\OrgMat$ by adding synthetic noise. 
Let $\mathrm{noise}(\OrgMat)$ be a random noise generator function that outputs a matrix of size $\nInsts \times \nAttrs$ based on the distribution of $\OrgMat$.
With the perturbation strength $\PerturbStrength \geq 0$, we generate a perturbed dataset with $\OrgMat^{(\MakeLowercase{\nRuns})} = \OrgMat + \PerturbStrength \cdot \mathrm{noise}(\mathbf{\OrgMat})$.
This design is conceptually similar to additive perturbations used in adversarial settings for neural networks~\cite{goodfellow2015explaining,fujiwara2025adversarial}.

We support multiple noise generators (e.g., uniform, Gaussian, and out-of-distribution variants) defined relative to the bounding box of $\OrgMat$ to capture common contamination patterns. 
By default, we use uniform noise as a simple, distribution-agnostic baseline that introduces unbiased off-manifold samples across the data space.

The range of $\PerturbStrength$ should be selected differently for each perturbation method to ensure plausibility.
For example, noise addition typically requires small values (e.g., $\PerturbStrength = 0.01$), as it directly perturbs feature values and can significantly alter each instance.
Similarly, outlier duplication uses relatively small strengths (e.g., $\PerturbStrength = 0.03$), corresponding to adding a small fraction of high-influence points.
In contrast, subsampling/resampling allows a broader range of strengths; for instance, $\PerturbStrength = 0.2$ corresponds to retaining 80\% of the data size, which remains a realistic scenario under limited observations.

\vspace{-1pt}
\subsection{Measuring Changes in Structural Preservation}
\label{sec:measure_changes}
\vspace{-1pt}

For each run, we use a panel of distortion measures to quantify structural preservation in a perturbed dataset and its corresponding low-dimensional representation.
The panel can be selected to capture complementary aspects of structural preservation.
For example, to span different local--global preservation ranges, we can include:
(i) local neighborhood preservation (e.g., \texttt{trustworthiness} and \texttt{continuity}~\cite{venna2001neighborhood}),
(ii) global-level preservation (e.g., \texttt{scale-normalized stress}~\cite{smelser2024normalized,smelser2026scale}), and
(iii) cluster-level structure (e.g., \texttt{steadiness} and \texttt{cohesiveness}~\cite{jeon2021measuring}).
When labels are available and aligned with the underlying structure, we can additionally compute label-aware measures~\cite{jeon2024classes} as complementary readouts.
When available, per-instance values from local and cluster-level measures support visualization and diagnosis (see \cref{sec:visualization}).

To avoid bias introduced by duplicated instances (e.g., in outlier duplication), we compute distortion measures only over the original set of instances.
For example, although outlier duplication increases the dataset size to $\nInsts + \lfloor \PerturbStrength \nInsts \rfloor$, we evaluate distortion measures using only the $\nInsts$ non-duplicated instances.
This ensures that sensitivity estimates reflect structural changes rather than artificial duplication effects.

All distortion measures are applied with fixed parameters, such as a consistent neighborhood size $k$ across runs, so that differences reflect perturbation effects and DR behavior rather than reconfiguration of the measures themselves. 
Each run yields a vector of distortion-measure values rather than a single value. 
This vector serves as the measurement layer for constructing sensitivity profiles in subsequent analysis.

When one-to-one instance correspondence is preserved, temporal stability measures could complement distortion measures by relating input-space changes to positional changes across layouts. 
Accordingly, for noise addition and each $\epsilon > 0$, we can additionally compute normalized temporal stress, temporal Shepard correlations~\cite{vernier2020quantitative}, and LCE after aligning each perturbed representation to the unperturbed one (e.g., with Procrustes transformation~\cite{gower2004procrustes}).
We cannot apply these measures to the other perturbations, which lack full correspondence.

\vspace{-1pt}
\subsection{Analysis Products and Visual Diagnosis}
\label{sec:visualization}
\vspace{-1pt}

As demonstrated in \cref{fig:swissroll_example}, our framework provides two primary visualizations for analyzing perturbation results: (1) shaded deviation line charts of sensitivity profiles (\cref{fig:swissroll_example}(a)) and (2) per-instance variation visualizations (\cref{fig:swissroll_example}(b)).

The shaded deviation line charts summarize how distortion-measure values change across perturbation strengths.
The mean value is shown as a line, and variability across runs is encoded as a shaded region representing the standard deviation.
In an initial design, we summarized all instances together; however, this can obscure localized effects of perturbations.
We compute statistics for the top-$A$\% of instances with the highest variability in their per-instance measure values. 
We refer to these instances as the \textit{high-variance subset}.
Variability is measured across perturbation strengths and runs, with \(A=10\) by default.

Plotting perturbation strength on the $x$-axis reveals the direction and rate of distortion-measure responses within each perturbation method. 
Sweeping strengths can expose nonlinear or nonmonotonic responses missed by evaluating a single condition (e.g., \cref{fig:swissroll_example}(a) and \cref{fig:tsne_vs_umap}(b)).

To identify perturbation-sensitive regions, we encode the variance of per-instance measure values using point size and the values themselves using color on the unperturbed representation. 
For example, \texttt{trustworthiness} colors indicate how well input data support neighborhoods in the representation, while point sizes indicate variation across perturbations.
These encodings collectively reveal regional differences in structural preservation and its variability, helping identify potentially unreliable patterns.

Juxtaposed sensitivity profiles (e.g., \cref{fig:tsne_vs_umap}(a)) support comparison across DR methods, datasets, and structural properties evaluated by different distortion measures.

\makeatletter
\def\thickhline{%
  \noalign{\ifnum0=`}\fi\hrule \@height \thickarrayrulewidth \futurelet
   \reserved@a\@xthickhline}
\def\@xthickhline{\ifx\reserved@a\thickhline
               \vskip\doublerulesep
               \vskip-\thickarrayrulewidth
             \fi
      \ifnum0=`{\fi}}
\makeatother

\newlength{\thickarrayrulewidth}
\setlength{\thickarrayrulewidth}{2\arrayrulewidth}

\subsection{Implementation and Parallelization}
We implement our framework in Python 3. 
Perturbation methods are implemented using NumPy/SciPy~\cite{virtanen2020scipy} and scikit-learn~\cite{pedregosa2011scikit} for random sampling, preprocessing, isolation forest-based outlier detection, and parallel computation. 
Distortion measures are computed using ZADU~\cite{jeon2023zadu} and visualizations are generated with matplotlib~\cite{hunter2007matplotlib}.

The pipeline is modular and supports repeated evaluation across perturbation strengths, DR methods, and distortion measures.
Most stages of the pipeline are embarrassingly parallel, including processing across perturbation methods, perturbation runs, DR methods, and datasets.
We use Joblib to parallelize these computations and reduce runtime.
\begin{table}[t]
\small
\centering
\caption{Summary of datasets used in the case studies.\vspace{-5pt}}
\label{tab:datasets}
\setlength{\tabcolsep}{2pt}
\renewcommand{\arraystretch}{0.9}
\begin{tabular}{p{1.2cm}rrp{5.9cm}}
\thickhline{}
Dataset & $\nInsts$ & $\nAttrs$ & Characteristics \\
\hline
\texttt{COIL-20} & 1,440 & 1,024 & Object images forming a 1D circular manifold\\ 
\\
\texttt{PBMC3k} & 2,700 & 2,000 &  Noisy single-cell data with heterogeneous cell populations\\
\texttt{C.\,elegans} & 6,188 & 50 & \setstretch{0.8}Trajectory-structured single-cell data with continuous progression and branching lineages\\
\texttt{MNIST} & 5,000 & 784 & Cluster-structured images with well-separated digit classes \\
\texttt{20 News.} & 5,000 & 1,000 & High-dim. sparse text data with overlapping semantic classes  \\
\thickhline{}
\end{tabular}
\vspace{-8pt}
\end{table}

\section{Case Studies}
\label{sec:case_studies}

We illustrate the utility of our framework through case studies using five datasets.
The case studies are organized around the three usage scenarios (cf.~\cref{sec:motivating_scenarios}).
Cases~1--3 present a connected sequence of analyses using the \texttt{C.\,elegans} dataset under outlier duplication, progressing from comparing PCA and MDS, through diagnosing their observed behavior, to using those findings to evaluate and design robustness-oriented variants.
Case~4 broadens these comparison and diagnostic analyses to the widely used nonlinear DR methods t-SNE and UMAP across five datasets and perturbation methods.
These selected DR methods represent linear, distance-based, and neighborhood-based DR formulations. 
The case studies demonstrate the analytical utility of the framework rather than provide an exhaustive benchmark or general ranking of DR methods.

\vspace{3pt}\noindent\textbf{Datasets.}
We use five datasets: (1) \texttt{COIL-20}~\cite{nene1996columbia}, (2) \texttt{PBMC3k}~\cite{10xgenomics,wolf2018scanpy}, (3) \texttt{C.\,elegans}~\cite{packer2019lineage,monocle3}, (4) \texttt{MNIST}~\cite{mnist}, and (5) \texttt{20~Newsgroups}~\cite{Lang95}.
These datasets span diverse structural characteristics, including circular manifolds, clusters, trajectories, and high-dimensional noisy data, as summarized in \cref{tab:datasets}. 
We use preprocessed versions of each dataset.
For \texttt{COIL-20}, background regions are removed from the images.
For \texttt{PBMC3k}, we use the dataset provided by the Scanpy library~\cite{wolf2018scanpy}.
For \texttt{C.\,elegans}, we follow preprocessing procedures described in the Monocle3 documentation~\cite{monocle3trajectories}.
For \texttt{MNIST}, we randomly subsample instances for computational efficiency. 
For \texttt{20~Newsgroups}, we first apply latent Dirichlet allocation~\cite{blei2003latent} to obtain a 1,000-dimensional representation, followed by random subsampling.
All features are standardized for \texttt{PBMC3k}, \texttt{C.\,elegans}, and \texttt{20~Newsgroups}.

\vspace{3pt}\noindent\textbf{Common settings.}
Unless otherwise stated, we use the following settings across the case studies.
We evaluate subsampling/resampling over perturbation strengths $\PerturbStrength \in [-0.1, 0.1]$, outlier duplication over $\PerturbStrength \in [0.0, 0.1]$, and noise addition over $\PerturbStrength \in [0.0, 0.02]$.
We use $\nRuns=10$ perturbation runs, which provides a practical balance between estimate stability and computational cost based on the convergence analysis in the supplementary materials.
The perturbation ranges were selected to represent subtle but observable data variations: at most a 10\% change in sample size, 10\% duplicated outliers, and a noise strength of 0.02. 
These ranges are not intended as universal thresholds and should be adapted to the plausible variations of the application dataset.
As distortion measures, we select \texttt{trustworthiness} \& \texttt{continuity}~\cite{venna2001neighborhood}, \texttt{steadiness} \& \texttt{cohesiveness}~\cite{jeon2021measuring}, and \texttt{scale-normalized stress}~\cite{smelser2024normalized,smelser2026scale}.
Also, in the supplementary materials, we test temporal stability measures used in the literature~\cite{vernier2020quantitative,wang2022joint}. 
We evaluate PCA, metric MDS, UMAP, and t-SNE across all the datasets, perturbation methods, and distortion measures. 
For each perturbed dataset, UMAP and t-SNE are repeated using five random seeds to capture algorithmic variability.
PCA is deterministic, whereas MDS is run with a fixed initialization and random state so that its variability arises from the data perturbation rather than random initialization.
We use the implementations available in scikit-learn~\cite{pedregosa2011scikit} and umap-learn~\cite{McInnes2018} with default hyperparameter settings. 

Below, we report representative findings that highlight key sensitivity behaviors of the evaluated DR methods.
We emphasize \texttt{trustworthiness} and \texttt{continuity} because they capture local neighborhood preservation and provide per-instance values that enable detailed visual diagnosis.
In contrast, \texttt{steadiness} and \texttt{cohesiveness} involve stochastic sampling in their computation, which can introduce additional variability not directly attributable to data perturbations.
Results across all selected distortion measures are provided in the supplementary materials.

\subsection{Case 1: Comparison of PCA and MDS}
\begin{figure}[t]
    \centering
    \includegraphics[
        alt={},
        width=\linewidth
    ]{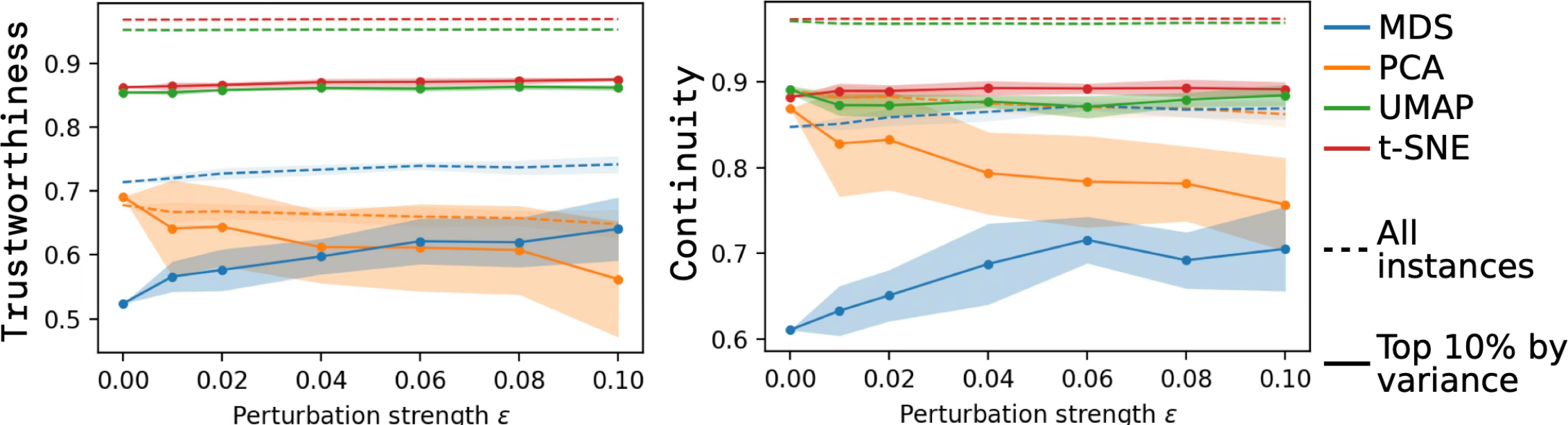}
    \caption{Sensitivity profiles (\texttt{trustworthiness}, \texttt{continuity}) under outlier duplication on the \texttt{C.\,elegans} dataset. 
    PCA exhibits a noticeable decrease in both measures, indicating degraded neighborhood preservation, while MDS surprisingly shows an increase, suggesting altered structural relationships under outlier influence.}
    \label{fig:celegans_trust_cont}
\end{figure}

This case study compares the sensitivity of PCA and MDS, corresponding to Scenario~1 in \cref{sec:motivating_scenarios}.
As a representative example, \cref{fig:celegans_trust_cont} shows \texttt{trustworthiness} and \texttt{continuity} values on the \texttt{C.\,elegans} dataset under data perturbation via outlier duplication.

Compared to MDS, UMAP, and t-SNE, PCA exhibits a clear degradation in local structure preservation, both for the high-variance subset and across all instances.
For example, at $\PerturbStrength = 0.1$, the mean \texttt{continuity} of the high-variance subset decreases from 0.87 to 0.76.
This behavior is consistent with prior findings that PCA is sensitive to outliers~\cite{ronchetti2009robust,xu2010robust,candes2011robust}.

In contrast, MDS shows an unexpected increase in both \texttt{trustworthiness} and \texttt{continuity}.
We hypothesize that this behavior arises from the interaction between the trajectory structure of the \texttt{C.\,elegans} dataset and the global distance-preserving objective of MDS.
The dataset contains continuous developmental trajectories, where certain regions may behave as structurally distinct or sparsely populated areas.
Since MDS optimizes pairwise distances across all instances, these regions may be underrepresented in the MDS result when few such instances exist.
By duplicating such instances, MDS may better preserve these local structures.

For the \texttt{C.\,elegans} dataset, MDS shows a higher overall \texttt{trustworthiness} value than PCA (see dashed lines), while exhibiting comparable or slightly lower \texttt{continuity}.
Moreover, the most sensitive instances (high-variance subset, solid lines) exhibit positive responses under outlier duplication. 
MDS also performs better according to \texttt{steadiness}, \texttt{cohesiveness}, and \texttt{scale-normalized stress} and exhibits greater stability under subsampling/resampling and noise addition perturbations (refer to the supplementary materials).

Under this evaluated setting, MDS 
performs better according to the selected distortion measures and exhibits smaller changes under perturbation than PCA.
However, these results alone do not establish MDS as the more suitable method for interpreting the neuron-development trajectories because the selected measures do not directly assess their visual distinguishability.
We investigate this issue in Case~2.

\begin{figure*}[t]
    \centering
    \includegraphics[
        alt={},
        width=0.95\linewidth
    ]{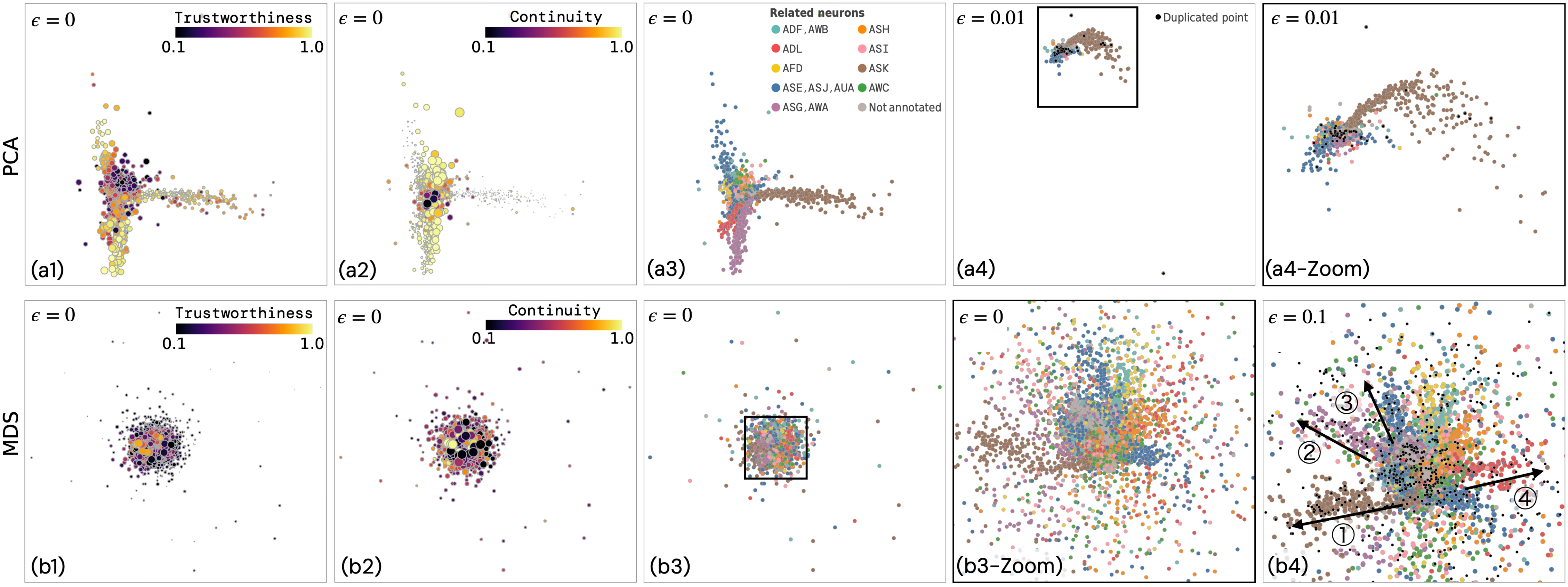}
    \caption{Detailed investigation of outlier duplication effects observed in \cref{fig:celegans_trust_cont}. 
    (a1--a3) show PCA representations colored by per-instance \texttt{trustworthiness}, \texttt{continuity}, and class label, respectively; in (a1) and (a2), point size encodes the variance of the corresponding per-instance measure values across perturbation strengths and runs.
    (b1--b3) show the corresponding visualizations for MDS. 
    (a4) and (b4) show embeddings at $\PerturbStrength=0.01$ (PCA) and $\PerturbStrength=0.1$ (MDS). 
    (a4-Zoom) and (b3-Zoom) provide zoomed-in views of the highlighted regions.
    }
    \label{fig:visual_investigation_pca_mds}
\end{figure*}

\subsection{Case 2: Understanding PCA and MDS Behaviors}

Building on Case 1, we further investigate how PCA and MDS respond to outlier duplication by examining their 2D representations under different perturbation strengths (PCA: $\PerturbStrength=0$ and $0.01$; MDS: $\PerturbStrength=0$ and $0.1$). 
This case corresponds to Scenario~2 in \cref{sec:motivating_scenarios}.

\vspace{3pt}\noindent\textbf{PCA.} In \cref{fig:visual_investigation_pca_mds}(a3), PCA reveals multiple trajectories corresponding to different neuron development paths, such as \texttt{ASK} (brown), \texttt{ASG,\,AWA} (purple), \texttt{ASE,\,ASJ,\,AUA} (blue), and \texttt{ADL} (red).
\cref{fig:visual_investigation_pca_mds}(a1) and (a2) show per-instance \texttt{trustworthiness} and \texttt{continuity} encoded by color, with value variability across perturbation strengths and runs encoded by point size.
We observe that regions associated with \texttt{ASK} exhibit relatively low variability, while other trajectories show higher variability, suggesting they are less stable under perturbation.
Consistent with this observation, even a small perturbation ($\PerturbStrength=0.01$, corresponding to 1\% duplicated outliers) significantly alters the PCA embedding, as shown in (a4).
In this case, only the \texttt{ASK} trajectory remains clearly visible.
Black points in (a4) indicate duplicated outliers.
The zoomed view (a4-Zoom) further confirms that other trajectories collapse or become indistinguishable from one another.
For this dataset, even a small amount of outlier duplication substantially alters the PCA representation and obscures several initially visible trajectories.

\vspace{3pt}\noindent\textbf{MDS.} \cref{fig:visual_investigation_pca_mds}(b1--b4) show the  results for MDS.
In (b1) and (b2), instances near the central region exhibit higher variability and lower \texttt{trustworthiness} and \texttt{continuity} values.
In contrast to PCA, the initial MDS embedding (b3-Zoom) primarily reveals the \texttt{ASK} trajectory.
However, as shown in (b4), introducing a stronger perturbation ($\PerturbStrength=0.1$) leads to the emergence of multiple trajectories corresponding to different neuron groups (\textcircled{\small 1}: \texttt{ASK}; \textcircled{\small 2}: \texttt{ASG,\,AWA}; \textcircled{\small 3}: \texttt{ASE,\,ASJ,\,AUA}; \textcircled{\small 4}: \texttt{ADL}).
These observations support our hypothesis from Case 1: by duplicating structurally important but underrepresented instances, MDS improves its ability to preserve trajectory-related local structures.

These visual diagnoses clarify the scope of the selected distortion measures. 
MDS's more favorable measure values indicate better preservation of the structural properties quantified by these measures.
Nevertheless, at $\epsilon=0$, several annotated developmental trajectories are more visually distinguishable in the PCA representation than in the MDS representation (\cref{fig:visual_investigation_pca_mds}(a3) and (b3)). 
This does not establish PCA as more faithful overall. 
Rather, the selected measures do not directly assess the task-specific visual distinguishability of these trajectories.

\begin{figure}[t]
    \centering
    \includegraphics[
        alt={},
        width=0.95\linewidth
    ]{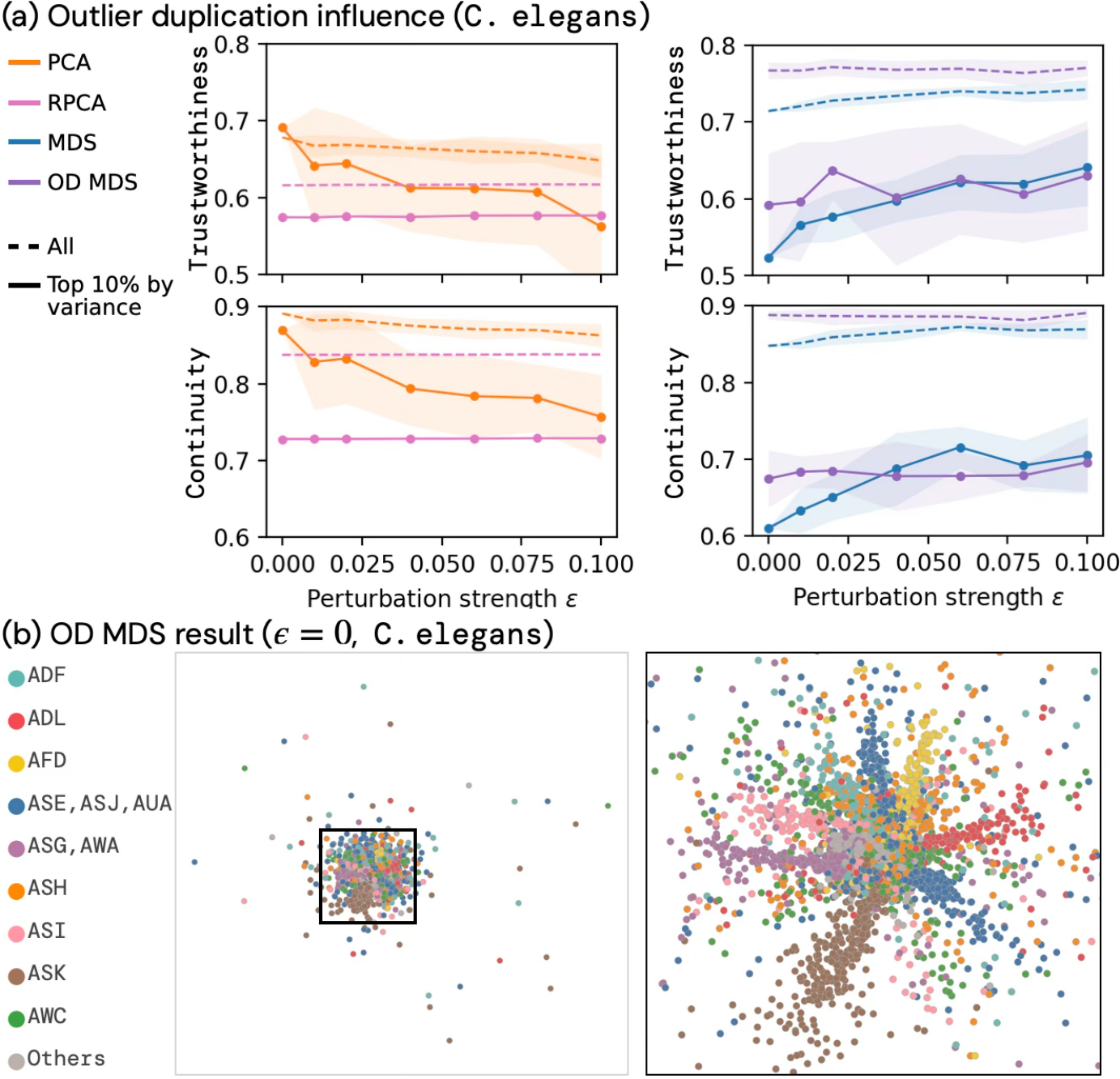}
    \caption{Evaluation of robust PCA and MDS variants. 
    (a) RPCA and OD MDS exhibit more stable distortion-measure values than their conventional counterparts. 
    (b) OD MDS shows clearer trajectories.}
    \label{fig:nonrobust_vs_robust}
\end{figure}

\subsection{Case 3. Design and Evaluation of Robust Methods}
\label{sec:case3}

Based on Cases~1 and 2, we examine how sensitivity analysis can support robustness-oriented method evaluation and development under outlier duplication (Scenario~3 in \cref{sec:motivating_scenarios}).
We demonstrate two uses: comparing PCA with an existing robust PCA (RPCA) method~\cite{candes2011robust,robust-pca-github} and developing a proof-of-concept MDS variant motivated by our earlier observations.
These examples illustrate the use of our framework rather than an exhaustive development of robust variants for all evaluated DR methods. Analogous extensions to other methods would require separate method-specific designs.

\vspace{3pt}\noindent\textbf{Outlier-duplication-enhanced MDS.} As shown in Case 1, \texttt{trustworthiness} and \texttt{continuity} of MDS increase when outliers are duplicated.
Motivated by this observation, we explicitly incorporate outlier duplication into the input of MDS.
We refer to this approach as outlier-duplication-enhanced MDS (OD MDS).
Specifically, instead of applying MDS directly to the perturbed dataset, we further augment the input by duplicating outlier instances using an additional strength parameter $\PerturbStrength'$, independent of the perturbation strength $\PerturbStrength$ used for sensitivity evaluation.
Based on the observation in \cref{fig:celegans_trust_cont} that MDS improves under relatively strong perturbation, we set $\PerturbStrength' = 0.5$ in this case study.

Formally, given a perturbed dataset $\smash{\OrgMat^{(r)} \in \mathbb{R}^{\nInsts' \times \nAttrs}}$, we construct an augmented dataset $\smash{\OrgMat^{(r)}_{\mathrm{aug}}}$ by duplicating outlier instances: $\smash{\OrgMat^{(r)}_{\mathrm{aug}} = [\OrgMat^{(r)};~ \widetilde{\OrgMat}^{(r)}]}$
where $\widetilde{\OrgMat}^{(r)}$ contains $\lfloor \PerturbStrength' \nInsts' \rfloor$ duplicated instances sampled with probability proportional to their anomaly scores.
We then apply MDS to $\smash{\OrgMat^{(r)}_{\mathrm{aug}}}$ to obtain the 2D representation.

\begin{figure*}[t]
    \centering
    \includegraphics[
        alt={},
        width=0.94\linewidth
    ]{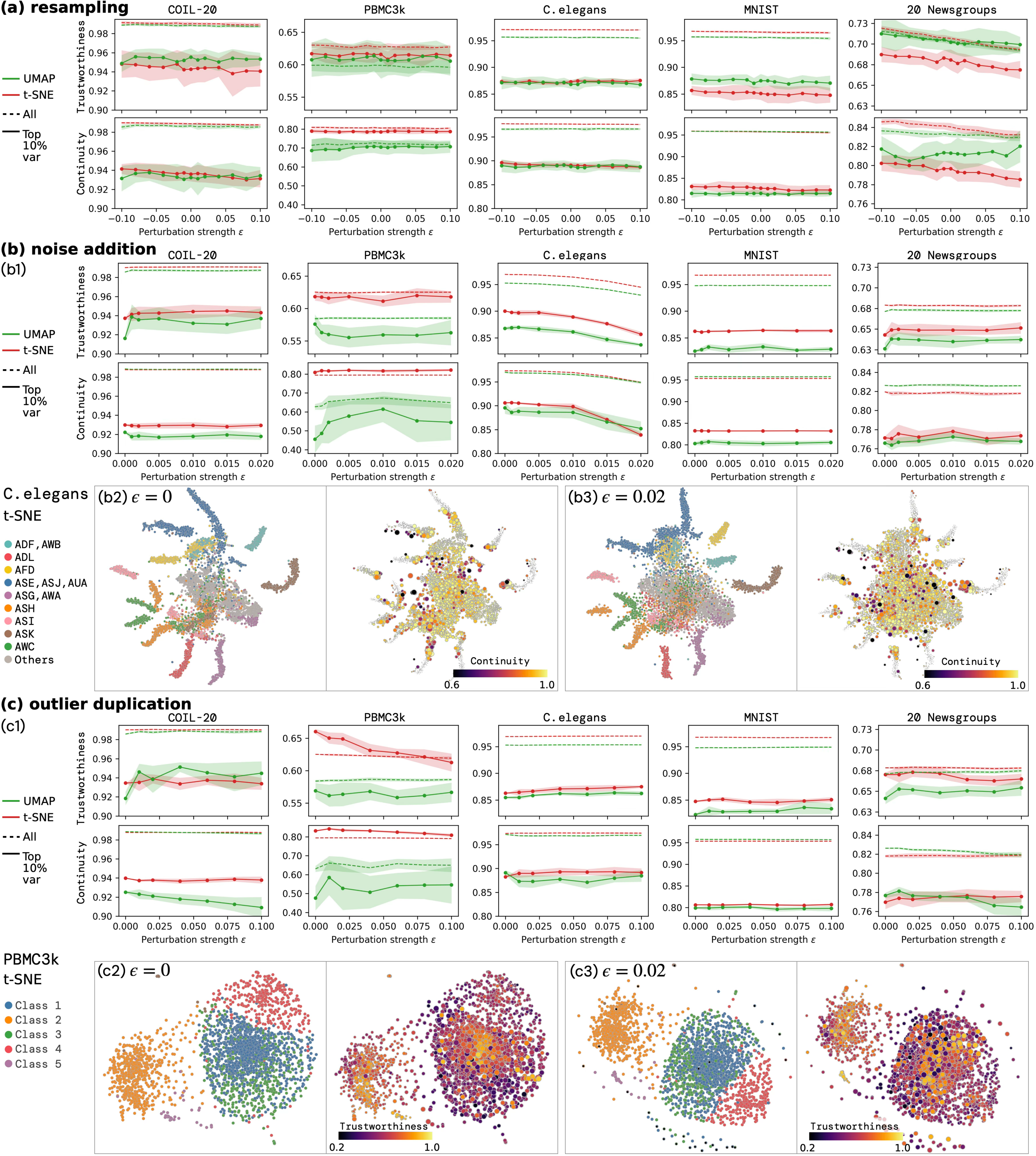}
    \caption{Sensitivity analysis of t-SNE and UMAP. 
    (a--c) show results for subsampling, noise addition, and outlier duplication, respectively. 
    t-SNE generally achieves higher \texttt{trustworthiness} and \texttt{continuity} values than UMAP, but both methods exhibit sensitivity to perturbations. 
    t-SNE representations (b2, b3, c2, c3) illustrate how noise alters trajectory structures (b2, b3) and how outlier duplication changes cluster structure (c2, c3).}
    \label{fig:tsne_vs_umap}
\end{figure*}

\vspace{3pt}\noindent\textbf{PCA vs.\ RPCA.}
The comparison between PCA and RPCA, as well as MDS and OD MDS, is shown in \cref{fig:nonrobust_vs_robust}(a).
On the \texttt{C.\,elegans} dataset, RPCA exhibits stable \texttt{trustworthiness} and \texttt{continuity} across perturbations.
However, its \texttt{trustworthiness} and \texttt{continuity} remain lower than those of PCA, indicating a trade-off between stability under perturbation and neighborhood preservation in this case.

\vspace{3pt}\noindent\textbf{MDS vs.\ OD MDS.} 
In contrast, OD MDS shows improved robustness compared to standard MDS.
Although variability across runs (shaded regions) remains noticeable, the mean values are more consistent across perturbation strengths for both all instances (dashed lines) and the high-variance subset (solid lines).
Moreover, OD MDS achieves higher mean values than MDS (e.g., \texttt{trustworthiness}: 0.77 vs.\ 0.71; \texttt{continuity}: 0.89 vs.\ 0.85 at $\PerturbStrength=0$).
To visually assess these improvements, \cref{fig:nonrobust_vs_robust}(b) shows a 2D representation at $\PerturbStrength=0$.
The zoomed view reveals clearer trajectory structures for multiple neuron groups, which are not apparent in the corresponding MDS result (\cref{fig:visual_investigation_pca_mds}(b3-Zoom)).

These results suggest that RPCA improves robustness at the cost of lower \texttt{trustworthiness} and \texttt{continuity}, whereas OD MDS improves both their values and consistency relative to MDS.

OD MDS is a proof-of-concept variant evaluated on \texttt{C.\,elegans} under outlier duplication. We have not compared OD MDS with controls that randomly select instances for duplication or adjust their influence through weighting in the MDS objective.
Consequently, these results do not establish whether anomaly-score-based instance selection is necessary for the observed improvement or whether alternative strategies could produce similar benefits. Its effectiveness on other datasets and under other
perturbation families remains untested.

\subsection{Case 4. Comparing and Diagnosing t-SNE and UMAP}

We compare and diagnose t-SNE and UMAP, two widely used nonlinear DR methods, following Scenarios~1 and 2.
\cref{fig:tsne_vs_umap} summarizes their \texttt{trustworthiness} and \texttt{continuity} under three perturbation methods across five datasets.

In many of the evaluated settings, t-SNE generally achieves higher \texttt{trustworthiness} and \texttt{continuity} values than UMAP (\cref{fig:tsne_vs_umap}(a--c)).
Similar trends are observed for other measures such as \texttt{steadiness}, \texttt{cohesiveness}, and \texttt{scale-normalized stress} (see the supplementary materials).
Under the evaluated datasets and configurations, t-SNE generally performs better according to these distortion measures.

However, sensitivity analysis reveals important nuances in how t-SNE behaves under perturbations.
For subsampling (\cref{fig:tsne_vs_umap}(a)), t-SNE exhibits a sharper decline in \texttt{trustworthiness} and \texttt{continuity} for \texttt{20~Newsgroups} as $\PerturbStrength$ increases.
For noise addition (\cref{fig:tsne_vs_umap}(b1)), both t-SNE and UMAP show substantial degradation in \texttt{trustworthiness} and \texttt{continuity} for \texttt{C.\,elegans}.
For outlier duplication (\cref{fig:tsne_vs_umap}(c1)), t-SNE shows a clear decrease in \texttt{trustworthiness} and \texttt{continuity} for \texttt{PBMC3k}, particularly for the high-variance subset.
Interestingly, for the \texttt{C.\,elegans} dataset, t-SNE exhibits slight improvements under outlier duplication, similar to MDS (cf.~\cref{fig:celegans_trust_cont}), suggesting that the outlier-duplication strategy introduced in \cref{sec:case3} may also benefit \mbox{t-SNE}.
To better understand these behaviors, we examine representative 2D representations.

For noise addition to the \texttt{C.\,elegans} dataset, the original embedding ($\PerturbStrength=0$, \cref{fig:tsne_vs_umap}(b2)) shows clear trajectory structures.
However, per-instance \texttt{continuity} values and their variance indicate that points near the center, where multiple neuron classes overlap, and along trajectory boundaries exhibit high variability.
This suggests that neighborhood relationships in these regions are unstable under perturbation.
After introducing small noise ($\PerturbStrength=0.02$, \cref{fig:tsne_vs_umap}(b3)), the central region expands, trajectories become less separated, and some structures collapse toward the center.
These changes indicate that certain visually salient patterns may be sensitive to noise and, therefore, less reliable.

For outlier duplication on \texttt{PBMC3k}, the original 2D representation ($\PerturbStrength=0$, \cref{fig:tsne_vs_umap}(c2)) shows two main clusters: one dominated by \texttt{Class\,2} (orange) and another mixing \texttt{Classes\,1} (blue), \texttt{3} (green), and \texttt{4} (red).
Variance in \texttt{trustworthiness} highlights instability in regions where \texttt{Class\,1} and \texttt{Class\,3} overlap.
After introducing a small number of duplicated outliers ($\PerturbStrength=0.02$, \cref{fig:tsne_vs_umap}(c3)), this mixed cluster becomes more condensed, and \texttt{Class\,1} points become more dispersed and intermixed with \texttt{Class\,3}.
This suggests that the apparent cluster structure in the original embedding is sensitive to outlier influence.

Overall, these results show that t-SNE, despite achieving high \texttt{trustworthiness} and \texttt{continuity} values, can exhibit sensitivity to different perturbations depending on the dataset structure.
In particular, it is sensitive to noise when revealing trajectory-like structures (e.g., \texttt{C.\,elegans}) and to outlier duplication in datasets with dense cluster regions (e.g., \texttt{PBMC3k}).
Such insights highlight the importance of sensitivity-aware analysis for interpreting DR visualizations.

Sec.~E of the supplementary materials presents a comparison with DynamicViz~\cite{sun2023dynamic} using the same bootstrap samples and resulting representations. 
UMAP exhibits substantially higher global and local distance variance than t-SNE, despite numerically similar mean \texttt{trustworthiness} across bootstrap samples. 
The highest-local-variance groups, selected separately for each DR method, also show different responses: \texttt{continuity} increases for t-SNE but decreases for UMAP relative to their respective unchanged-input references. 
These findings illustrate that distance variability across representations alone does not indicate the level of structural preservation or whether it improves or deteriorates under resampling.
Our framework additionally examines how structural preservation responds to different perturbation families and strengths.

\section{Discussion}
\textbf{Role and generalizability of sensitivity analysis.}
Our results show that perturbation-driven sensitivity analysis provides information about DR behavior that one-shot distortion measures alone do not capture. 
In particular, it reveals whether a DR method that appears effective on an unperturbed dataset continues to preserve relevant structural relationships under plausible variations of the input. 
This analysis is most relevant when such variations represent nuisance conditions, such as sampling variability, observational noise, or outlier contamination. 
When input changes represent meaningful temporal evolution, the representation should reflect them, making cross-layout stability a separate evaluation objective, as discussed in \cref{sec:dynamic_dr}.

Sensitivity reflects interactions among the dataset, DR method, configuration, perturbation, and distortion measure. Weak cluster separation, continuous structure, noise, or rare observations may increase sensitivity.
Methods may respond differently to the same perturbation, and measures may reveal different aspects of change. Sensitivity profiles therefore support comparisons within evaluated settings rather than universal classifications of methods as stable or unstable. 
Diagnosing affected structures may suggest broader explanations, but establishing their generality requires additional datasets and configurations.

The in-depth analyses in Cases~1--3 focus on \texttt{C.\,elegans} under outlier duplication. 
Although Case~4 and Sec.~A provide broader comparisons, further evaluation is needed to determine whether the explanations and robustness improvements explored in Cases 1--3 generalize to other datasets and perturbation families.
Also, although designed to be model-agnostic, the framework remains empirically untested on neural-network-based and other nonlinear parametric DR methods~\cite{sainburg2021parametric}.

Our supplementary analysis shows that temporal stability and distortion measures can yield different assessments because they evaluate cross-layout positional coherence and within-representation structural preservation, respectively. 
Designed for temporally coordinated layouts, temporal measures require instance correspondence, limiting their use here to noise addition. 
Global alignment also cannot resolve nonrigid rearrangements between independently generated representations. 
Since positional changes may appropriately reflect input changes, they alone do not establish deterioration in structural preservation. 
Temporal measures can therefore complement our framework when correspondence is preserved but cannot replace distortion-based sensitivity analysis.

\vspace{3pt}\noindent\textbf{Interpretation of sensitivity results.}
A key limitation is that the framework does not prescribe a universal threshold for when a change becomes unacceptable. 
The practical meaning of degradation depends on the analytic task and on which structural properties matter for that task. 
For example, a mild change in a global measure may be tolerable for exploratory overview, whereas a similar change in a neighborhood- or cluster-level measure may be critical for tasks such as cluster interpretation or anomaly detection. 
We therefore position the framework as supporting calibrated interpretation rather than binary judgments of ``stable'' versus ``unstable.'' 
The case studies illustrate how per-instance diagnostics can guide qualitative investigation of changes in clusters and trajectories. These diagnostics quantify neighborhood-preservation variability. Thus, they do not directly measure cluster separation or trajectory-shape preservation. Quantitatively validating their ability to identify such structural changes remains future work.

\vspace{3pt}\noindent\textbf{Guidance for balancing structural preservation and sensitivity.}
Analysts should identify plausible perturbations, application-relevant strength ranges, and task-relevant distortion measures, then examine both the resulting measure values and their variation. 
For example, under outlier duplication on \texttt{PBMC3k} (\cref{fig:tsne_vs_umap}(c1)), t-SNE's \texttt{trustworthiness} decreases but remains higher than UMAP's throughout the evaluated range. 
This favors t-SNE for neighborhood preservation as measured by \texttt{trustworthiness}, but does not establish that the structures supporting the intended analysis remain consistent.
Its greater sensitivity suggests a need for closer inspection using the diagnostic visualizations. 
When profiles cross, or measures disagree, method selection depends on the perturbation conditions and structural properties relevant to the task. 
The framework therefore supports conditional recommendations rather than a universal ranking or threshold for acceptable change.

\vspace{3pt}\noindent\textbf{Computational cost.} 
For each perturbation type, the framework generates $R \times T$ perturbed datasets and performs $R \times T \times S$ embedding runs, where $T$ is the number of strengths and $S$ the number of DR seeds, followed by distortion-measure evaluation. 
Because these runs are independent, they can be parallelized.
Evaluating four DR methods and five measures with $R=10$ and $T\approx10$ was practical in our case studies. 
For the \texttt{C.\,elegans} dataset, noise addition and outlier duplication ($T=7$) each took under an hour (43 minutes on average) on a 2024 MacBook Pro (M4 Max, 16 cores, 128 GB).

In practice, analysts can initially test a few strengths and measures, then refine conditions showing notable changes. 
GPU-accelerated DR~\cite{raschka2020machine} can reduce runtime, but measure computation may remain a bottleneck even with optimized implementations such as tf-projection-qm~\cite{tensorpqm}, motivating more efficient and scalable implementations.

\vspace{3pt}\noindent\textbf{Perturbation and distortion measure design for sensitivity analysis.}
As a first step, we introduced three perturbation methods to capture common sources of data variation, but they are not exhaustive.
Future work could consider other conditions, such as out-of-distribution samples that substantially shift the data distribution and resulting representations.

Our framework uses existing distortion measures to quantify changes across perturbations, addressing the challenge of directly comparing multiple low-dimensional representations.
However, developing sensitivity-specific measures that explicitly capture changes in local, cluster, and global structure could further improve analysis.
Even without new measures, systematically evaluating existing distortion measures across datasets and DR methods may reveal which measures are most informative for sensitivity analysis.

\section{Conclusion}
We presented a perturbation-driven framework that extends one-shot DR evaluation by tracking structural preservation, quantified using distortion measures, under controlled, plausible input variations. 
Our case studies show that methods that perform well according to distortion measures can remain sensitive to perturbations, with instability in visually salient trajectories and clusters, while some methods improve structural preservation under particular perturbations. 
Sensitivity thus reflects interactions between the dataset, DR method, and perturbation. 
By combining sensitivity profiles with diagnostic visualizations, the framework supports informed method selection, interpretation of visual patterns, and robustness-oriented method development.

Future work includes richer perturbation designs, evaluation of additional DR families, sensitivity-oriented measures, and robustness-aware DR objectives. 
More broadly, our findings encourage evaluating both how well a representation preserves structure in an observed dataset and how reliably that structure persists under plausible data variation.

\acknowledgments{
  This work has been supported in part by the National Research Foundation of Korea (NRF) grant funded by the Korea government (MSIT) (RS-2026-25605977).
}

\bibliographystyle{abbrv-doi-hyperref}

\bibliography{template}

@inbook{jolliffe1986principal,
  title={Principal Component Analysis and Factor Analysis},
  author={Jolliffe, Ian T},
  booktitle={Principal Component Analysis},
  pages={115--128},
  year={1986},
  publisher={Springer},
  doi       = {10.1007/978-1-4757-1904-8_7}
}

@article{torgerson1952,
  author={Torgerson, Warren S.},
  title={Multidimensional scaling: I. Theory and method},
  journal={Psychometrika},
  year={1952},
  volume={17},
  number={4},
  pages={401--419},
  doi     = {10.1007/BF02288916}
}

@article{hotelling1933analysis,
  title={Analysis of a complex of statistical variables into principal components},
  author={Hotelling, Harold},
  journal={J. Educ. Psychol.},
  volume={24},
  number={6},
  pages={417},
  year={1933},
  publisher={Warwick \& York},
  doi     = {10.1037/h0071325}
}

@article{bertini2011quality,
  title={Quality metrics in high-dimensional data visualization: An overview and systematization},
  author={Bertini, Enrico and Tatu, Andrada and Keim, Daniel},
  journal={IEEE Trans. Vis. Comput. Graph.},
  volume={17},
  number={12},
  pages={2203--2212},
  year={2011},
  publisher={IEEE},
  doi       = {10.1109/TVCG.2011.229}
}

@article{mcinnes2018umap,
  title={{UMAP}: Uniform manifold approximation and projection for dimension reduction},
  author={McInnes, Leland and Healy, John and Melville, James},
  journal={arXiv:1802.03426},
  year={2018},
  doi={https://doi.org/10.48550/arXiv.1802.03426}
}

@article{espadoto2021toward,
  author={M. {Espadoto} and R. M. {Martins} and A. {Kerren} and N. S. T. {Hirata} and A. C. {Telea}},
  journal={IEEE Trans. Vis. Comput. Graph.}, 
  title={Toward a Quantitative Survey of Dimension Reduction Techniques}, 
  year={2021},
  volume={27},
  number={3},
  pages={2153-2173},
  doi     = {10.1109/TVCG.2019.2944182}
 }

@article{cunningham2015linear,
  title={Linear dimensionality reduction: Survey, insights, and generalizations},
  author={Cunningham, John P and Ghahramani, Zoubin},
  journal={J. Mach. Learn. Res.},
  volume={16},
  number={1},
  pages={2859--2900},
  year={2015},
  publisher={JMLR. org},
  note={\url{https://jmlr.org/papers/volume16/cunningham15a/cunningham15a.pdf}},
}

@techreport{van2009dimensionality,
  author={van der Maaten, Laurens and Postma, Eric and van den Herik, Jaap},
  title={Dimensionality reduction: A comparative review},
  institution={Tilburg University Technical Report},
  year={2009},
  number={TiCC-TR 2009-005},
  note={36 pages, \url{https://lvdmaaten.github.io/publications/papers/TR_Dimensionality_Reduction_Review_2009.pdf}}
}

@inproceedings{garcia2013stability,
  title={Stability comparison of dimensionality reduction techniques attending to data and parameter variations},
  author={Garc{\'\i}a Fern{\'a}ndez, Francisco Javier and Verleysen, Michel and Lee, John A and D{\'\i}az Blanco, Ignacio and others},
  booktitle={Proc. VAMP},
  year={2013},
  organization={Eurographics},
  doi={10.2312/PE.VAMP.VAMP2013.005-009},
  note={5 pages}
}

@article{kruskal1964multidimensional,
  title={Multidimensional scaling by optimizing goodness of fit to a nonmetric hypothesis},
  author={Kruskal, Joseph B},
  journal={Psychometrika},
  volume={29},
  number={1},
  pages={1--27},
  year={1964},
  publisher={Springer-Verlag},
  doi       = {10.1007/BF02289565}
}

@inproceedings{hyeon2025unveiling,
author = {Jeon, Hyeon and Lee, Hyunwook and Kuo, Yun-Hsin and Yang, Taehyun and Archambault, Daniel and Ko, Sungahn and Fujiwara, Takanori and Ma, Kwan-Liu and Seo, Jinwook},
title = {Unveiling High-dimensional Backstage: A Survey for Reliable Visual Analytics with Dimensionality Reduction},
booktitle = {Proc. CHI},
year = {2025},
isbn = {9798400713941},
publisher = {ACM},
doi = {10.1145/3706598.3713551},
articleno = {394},
numpages = {24}
}

@article{liu2025assessing,
  title={Assessing and improving reliability of neighbor embedding methods: A map-continuity perspective},
  author={Liu, Zhexuan and Ma, Rong and Zhong, Yiqiao},
  journal={Nat. Commun.},
  volume={16},
  number={1},
  pages={5037},
  year={2025},
  publisher={Nature Publishing Group UK London},
  doi       = {10.1038/s41467-025-60434-9}
}

@inproceedings{hamad2018stability,
  title={Stability analysis of the {t-SNE} algorithm for human activity pattern data},
  author={Hamad, Rebeen Ali and J{\"a}rpe, Eric and Lundstr{\"o}m, Jens},
  booktitle={Proc. SMC},
  pages={1839--1845},
  year={2018},
  organization={IEEE},
  doi          = {10.1109/SMC.2018.00318}
}

@article{liu2016visualizing,
  title={Visualizing high-dimensional data: Advances in the past decade},
  author={Liu, Shusen and Maljovec, Dan and Wang, Bei and Bremer, Peer-Timo and Pascucci, Valerio},
  journal={IEEE Trans. Vis. Comput. Graph.},
  volume={23},
  number={3},
  pages={1249--1268},
  year={2016},
  publisher={IEEE},
  doi       = {10.1109/TVCG.2016.2640960}
}

@article{cashman2025critical,
  title={A critical analysis of the usage of dimensionality reduction in four domains},
  author={Cashman, Dylan and Keller, Mark and Jeon, Hyeon and Kwon, Bum Chul and Wang, Qianwen},
  journal={IEEE Trans. Vis. Comput. Graph.},
  year={2025},
  publisher={IEEE},
  volume={31},
  number={10},
  pages={9405-9423},
  doi = {10.1109/TVCG.2025.3567989}
}

@article{hullman2018pursuit,
  title={In pursuit of error: A survey of uncertainty visualization evaluation},
  author={Hullman, Jessica and Qiao, Xiaoli and Correll, Michael and Kale, Alex and Kay, Matthew},
  journal={IEEE Trans. Vis. Comput. Graph.},
  volume={25},
  number={1},
  pages={903--913},
  year={2019},
  publisher={IEEE},
  doi       = {10.1109/TVCG.2018.2864889}
}

@inproceedings{jeon2023zadu,
  title={{ZADU}: A python library for evaluating the reliability of dimensionality reduction embeddings},
  author={Jeon, Hyeon and Cho, Aeri and Jang, Jinhwa and Lee, Soohyun and Hyun, Jake and Ko, Hyung-Kwon and Jo, Jaemin and Seo, Jinwook},
  booktitle={Proc. VIS},
  pages={196--200},
  year={2023},
  organization={IEEE},
  doi          = {10.1109/VIS54172.2023.00048}
}

@article{sacha2015role,
  title={The role of uncertainty, awareness, and trust in visual analytics},
  author={Sacha, Dominik and Senaratne, Hansi and Kwon, Bum Chul and Ellis, Geoffrey and Keim, Daniel A},
  journal={IEEE Trans. Vis. Comput. Graph.},
  volume={22},
  number={1},
  pages={240--249},
  year={2016},
  publisher={IEEE},
  doi       = {10.1109/TVCG.2015.2467591}
}

@article{machado2025necessary,
author = {Machado, A. and Behrisch, M. and Telea, A.},
title = {Necessary but not Sufficient: Limitations of Projection Quality Metrics},
journal = {Comput. Graph. Forum},
volume = {44},
number = {3},
pages = {e70101},
doi = {10.1111/cgf.70101},
year = {2025}
}

@article{jeon2021measuring,
  title={Measuring and explaining the inter-cluster reliability of multidimensional projections},
  author={Jeon, Hyeon and Ko, Hyung-Kwon and Jo, Jaemin and Kim, Youngtaek and Seo, Jinwook},
  journal={IEEE Trans. Vis. Comput. Graph.},
  volume={28},
  number={1},
  pages={551--561},
  year={2022},
  publisher={IEEE},
  doi       = {10.1109/TVCG.2021.3114833}
}

@ARTICLE{jeon2024classes,
  author={Jeon, Hyeon and Kuo, Yun-Hsin and Aupetit, Michaël and Ma, Kwan-Liu and Seo, Jinwook},
  journal={IEEE Trans. Vis. Comput. Graph.}, 
  title={Classes are Not Clusters: Improving Label-Based Evaluation of Dimensionality Reduction}, 
  year={2024},
  volume={30},
  number={1},
  pages={781-791},
  doi={10.1109/TVCG.2023.3327187}
}

@article{venna2006local,
  title={Local multidimensional scaling},
  author={Venna, Jarkko and Kaski, Samuel},
  journal={Neural Netw.},
  volume={19},
  number={6-7},
  pages={889--899},
  year={2006},
  publisher={Elsevier},
  doi       = {10.1016/j.neunet.2006.05.014}
}

@inproceedings{venna2001neighborhood,
  title={Neighborhood preservation in nonlinear projection methods: An experimental study},
  author={Venna, Jarkko and Kaski, Samuel},
  booktitle={Proc. ICANN},
  pages={485--491},
  year={2001},
  organization={Springer},
  doi          = {10.1007/3-540-44668-0_68}
}

@article{sun2023dynamic,
  title={Dynamic visualization of high-dimensional data},
  author={Sun, Eric D and Ma, Rong and Zou, James},
  journal={Nat. Comp. Sci.},
  volume={3},
  number={1},
  pages={86--100},
  year={2023},
  publisher={Nature Publishing Group US New York},
  doi       = {10.1038/s43588-022-00380-4}
  
}

@article{kobak2021initialization,
  title={Initialization is critical for preserving global data structure in both {t-SNE} and {UMAP}},
  author={Kobak, Dmitry and Linderman, George C},
  journal={Nat. Biotechnol.},
  volume={39},
  number={2},
  pages={156--157},
  year={2021},
  publisher={Nature Publishing Group US New York},
  doi={https://doi.org/10.1038/s41587-020-00809-z}
}

@article{jung2025ghostumap2,
  title={{GhostUMAP2}: Measuring and Analyzing $(r, d)$-Stability of {UMAP}},
  author={Jung, Myeongwon and Fujiwara, Takanori and Jo, Jaemin},
  journal={IEEE Trans. Vis. Comput. Graph.},
  publisher={IEEE},
  year={2026},
  volume={32},
  number={1},
  pages={353-362},
  doi       = {10.1109/TVCG.2025.3633894}
  
}

@inproceedings{jung2024ghostumap,
  title={{GhostUMAP}: Measuring Pointwise Instability in Dimensionality Reduction},
  author={Jung, Myeongwon and Fujiwara, Takanori and Jo, Jaemin},
  booktitle={Proc. VIS},
  pages={161--165},
  year={2024},
  organization={IEEE},
  doi          = {10.1109/VIS55277.2024.00040}
}

@article{lehmann2015optimal,
  title={Optimal sets of projections of high-dimensional data},
  author={Lehmann, Dirk J and Theisel, Holger},
  journal={IEEE Trans. Vis. Comput. Graph.},
  volume={22},
  number={1},
  pages={609--618},
  year={2016},
  publisher={IEEE},
  doi       = {10.1109/TVCG.2015.2467132}
}

@article{wattenberg2016use,
  title={How to use {t-SNE} effectively},
  author={Wattenberg, Martin and Vi{\'e}gas, Fernanda and Johnson, Ian},
  journal={Distill},
  volume={1},
  number={10},
  pages={e2},
  year={2016},
  doi     = {10.23915/distill.00002}
}

@article{sips2009selecting,
  title={Selecting good views of high-dimensional data using class consistency},
  author={Sips, Mike and Neubert, Boris and Lewis, John P and Hanrahan, Pat},
  journal={Comput. Graph. Forum},
  volume={28},
  number={3},
  pages={831--838},
  year={2009},
  organization={Wiley Online Library},
  doi          = {10.1111/j.1467-8659.2009.01467.x}
}

@article{lee2009quality,
  title={Quality assessment of dimensionality reduction: Rank-based criteria},
  author={Lee, John A and Verleysen, Michel},
  journal={Neurocomputing},
  volume={72},
  number={7-9},
  pages={1431--1443},
  year={2009},
  publisher={Elsevier},
  doi       = {10.1016/j.neucom.2008.12.017}
}

@inproceedings{fujiwara2023feature,
  title={Feature learning for nonlinear dimensionality reduction toward maximal extraction of hidden patterns},
  author={Fujiwara, Takanori and Kuo, Yun-Hsin and Ynnerman, Anders and Ma, Kwan-Liu},
  booktitle={Proc. PacificVis},
  pages={122--131},
  year={2023},
  organization={IEEE},
  doi          = {10.1109/PacificVis56936.2023.00021}
}

@article{serneels2008principal,
  title={Principal component analysis for data containing outliers and missing elements},
  author={Serneels, Sven and Verdonck, Tim},
  journal={Comput. Stat. Data. Anal.},
  volume={52},
  number={3},
  pages={1712--1727},
  year={2008},
  publisher={Elsevier},
  doi       = {10.1016/j.csda.2007.05.024}
}

@article{sibson1979studies,
  title={Studies in the robustness of multidimensional scaling: Perturbational analysis of classical scaling},
  author={Sibson, Robin},
  journal={J. R. Stat. Soc. Ser. B Methodol.},
  volume={41},
  number={2},
  pages={217--229},
  year={1979},
  publisher={Wiley Online Library},
  doi       = {10.1111/j.2517-6161.1979.tb01076.x}
}

@article{green1975robustness,
  title={On the robustness of multidimensional scaling techniques},
  author={Green, Paul E},
  journal={J. Mark. Res.},
  volume={12},
  number={1},
  pages={73--81},
  year={1975},
  publisher={SAGE Publications Sage CA: Los Angeles, CA},
  doi       = {10.1177/002224377501200111}
}

@article{zabel2025visualizing,
  title={Visualizing stability: A sensitivity analysis framework for {t-SNE} embeddings},
  author={Zabel, Susanne and Hennig, Philipp and Nieselt, Kay},
  journal={Front. Bioinform.},
  volume={5},
  pages={1719516},
  year={2025},
  publisher={Frontiers Media SA},
  doi       = {10.3389/fbinf.2025.1719516}
}

@article{johnson2022embedr,
  title={{EMBEDR}: Distinguishing signal from noise in single-cell omics data},
  author={Johnson, Eric M and Kath, William and Mani, Madhav},
  journal={Patterns},
  volume={3},
  number={3},
  year={2022},
  publisher={Elsevier},
  doi       = {10.1016/j.patter.2022.100443}
}

@article{xia2024statistical,
  title={Statistical method {scDEED} for detecting dubious {2D} single-cell embeddings and optimizing {t-SNE} and {UMAP} hyperparameters},
  author={Xia, Lucy and Lee, Christy and Li, Jingyi Jessica},
  journal={Nat. Commun.},
  volume={15},
  number={1},
  pages={1753},
  year={2024},
  publisher={Nature Publishing Group UK London},
  doi       = {10.1038/s41467-024-45891-y}
}

@article{pang1997approaches,
  title={Approaches to uncertainty visualization},
  author={Pang, Alex T and Wittenbrink, Craig M and Lodha, Suresh K and others},
  journal={Vis. Comput.},
  volume={13},
  number={8},
  pages={370--390},
  year={1997},
  doi     = {10.1007/s003710050111}
}

@inproceedings{potter2011quantification,
  title={From quantification to visualization: A taxonomy of uncertainty visualization approaches},
  author={Potter, Kristin and Rosen, Paul and Johnson, Chris R},
  booktitle={Proc. WoCoUQ},
  pages={226--249},
  year={2011},
  organization={Springer},
  doi          = {10.1007/978-3-642-32677-6_15}
}

@article{nonato2018multidimensional,
  title={Multidimensional projection for visual analytics: Linking techniques with distortions, tasks, and layout enrichment},
  author={Nonato, Luis Gustavo and Aupetit, Michael},
  journal={IEEE Trans. Vis. Comput. Graph.},
  volume={25},
  number={8},
  pages={2650--2673},
  year={2018},
  publisher={IEEE},
  doi       = {10.1109/TVCG.2018.2846735}
}

@article{van2008visualizing,
  title={Visualizing data using {t-SNE}},
  author={van der Maaten, Laurens and Hinton, Geoffrey},
  journal={J. Mach. Learn. Res.},
  volume={9},
  number={11},
  year={2008},
  note={\url{https://www.jmlr.org/papers/volume9/vandermaaten08a/vandermaaten08a.pdf}}
}

@article{huertas2023exploring,
  title={Exploring dimensionality reduction techniques in multilingual transformers},
  author={Huertas-Garc{\'\i}a, {\'A}lvaro and Mart{\'\i}n, Alejandro and Huertas-Tato, Javier and Camacho, David},
  journal={Cogn. Comput.},
  volume={15},
  number={2},
  pages={590--612},
  year={2023},
  publisher={Springer},
  doi={10.1007/s12559-022-10066-8}
}

@article{xuan2025attributionscanner,
  title={{AttributionScanner}: A visual analytics system for model validation with metadata-free slice finding},
  author={Xuan, Xiwei and Ono, Jorge Piazentin and Gou, Liang and Ma, Kwan-Liu and Ren, Liu},
  journal={IEEE Trans. Vis. Comput. Graph.},
  year={2025},
  volume={31},
  issue={10},
  pages={7436--7447},
  doi={10.1109/TVCG.2025.3546644}
}

@article{rousseeuw2011robust,
  title={Robust statistics for outlier detection},
  author={Rousseeuw, Peter J and Hubert, Mia},
  journal={Wiley Interdisciplinary Reviews: Data Mining and Knowledge Discovery},
  volume={1},
  number={1},
  pages={73--79},
  year={2011},
  publisher={Wiley Online Library},
  doi       = {10.1002/widm.2}
}

@book{wasserman2004all,
  title={All of Statistics: A Concise Course in Statistical Inference},
  author={Wasserman, Larry},
  volume={26},
  year={2004},
  publisher={Springer},
  doi       = {10.1007/978-0-387-21736-9}
}

@book{fuller2009measurement,
  title={Measurement Error Models},
  author={Fuller, Wayne A},
  year={2009},
  publisher={John Wiley \& Sons},
  doi       = {10.1002/9780470316665}
}

@article{kamal2021recent,
  title={Recent advances and challenges in uncertainty visualization: A survey},
  author={Kamal, Aasim and Dhakal, Parashar and Javaid, Ahmad Y and Devabhaktuni, Vijay K and Kaur, Devinder and Zaientz, Jack and Marinier, Robert},
  journal={J. Vis.},
  volume={24},
  number={5},
  pages={861--890},
  year={2021},
  publisher={Springer},
  doi       = {10.1007/s12650-021-00755-1}
}

@INPROCEEDINGS{liu2008isolation,
  author={Liu, Fei Tony and Ting, Kai Ming and Zhou, Zhi-Hua},
  booktitle={Proc. ICDM},
  title={Isolation Forest}, 
  year={2008},
  volume={},
  number={},
  pages={413-422},
  doi={10.1109/ICDM.2008.17}
}

@article{virtanen2020scipy,
  author={Virtanen, Pauli and Gommers, Ralf and Oliphant, Travis E and Haberland, Matt and  Reddy, Tyler and others},
  title   = {{SciPy} 1.0: Fundamental Algorithms for Scientific
            Computing in {Python}},
  journal = {Nat. Methods},
  year    = {2020},
  volume  = {17},
  pages   = {261--272},
  doi={10.1038/s41592-019-0686-2}
}

@article{pedregosa2011scikit,
  title={Scikit-learn: Machine Learning in {P}ython},
  author={Pedregosa, Fabian and Varoquaux, Ga{\"e}l and Gramfort, Alexandre and Michel, Vincent and Thirion, Bertrand and others},
  journal={J. Mach. Learn. Res.},
  volume={12},
  pages={2825--2830},
  year={2011},
  doi={10.48550/arXiv.1201.0490}
}

@Article{hunter2007matplotlib,
  Author    = {Hunter, J. D.},
  Title     = {Matplotlib: A {2D} graphics environment},
  Journal   = {Comput. Sci. Eng.},
  Volume    = {9},
  Number    = {3},
  Pages     = {90--95},
  publisher = {IEEE COMPUTER SOC},
  year      = 2007,
  doi = {10.1109/MCSE.2007.55}
}

@article{tenenbaum2000global,
  title={A global geometric framework for nonlinear dimensionality reduction},
  author={Tenenbaum, Joshua B and Silva, Vin de and Langford, John C},
  journal={Science},
  volume={290},
  number={5500},
  pages={2319--2323},
  year={2000},
  publisher={American Association for the Advancement of Science},
  doi       = {10.1126/science.290.5500.2319}
}

@inproceedings{goodfellow2015explaining,
  author   ={Ian J. Goodfellow and
               Jonathon Shlens and
               Christian Szegedy},
  title    ={Explaining and Harnessing Adversarial Examples},
  booktitle={Proc. ICLR},
  year     ={2015},
  doi = {https://doi.org/10.48550/arXiv.1412.6572}
}

@inproceedings{smelser2024normalized,
  title={``{Normalized} Stress''' is Not Normalized: How to Interpret Stress Correctly},
  author={Smelser, Kiran and Miller, Jacob and Kobourov, Stephen},
  booktitle={Proc. BELIV},
  pages={41--50},
  year={2024},
  organization={IEEE},
  doi          = {10.1109/BELIV64461.2024.00010}
}

@techreport{nene1996columbia,
  title={{C}olumbia {O}bject {I}mage {L}ibrary ({COIL-20})},
  author={Nene, Sameer A and Nayar, Shree K and Murase, Hiroshi and others},
  year={1996},
  institution={CUCS-006-96},
  note={\url{https://cave.cs.columbia.edu/repository/COIL-20}}
}

@article{wolf2018scanpy,
  title={Scanpy: Large-scale single-cell gene expression data analysis},
  author={Wolf, F. Alexander and Angerer, Philipp and Theis, Fabian J.},
  journal={Genome Biol.},
  year={2018},
  doi     = {10.1186/s13059-017-1382-0}
}

@misc{10xgenomics,
  title={Datasets},
  author={{10x Genomics}},
  year={2025},
  howpublished={\url{https://www.10xgenomics.com/datasets}},
  note={Accessed: 2026-03-27}
}

@article{packer2019lineage,
  title={A lineage-resolved molecular atlas of \textit{C. elegans} embryogenesis at single-cell resolution},
  author={Packer, Jonathan S and Zhu, Qin and Huynh, Chau and Sivaramakrishnan, Priya and Preston, Elicia and others},
  journal={Science},
  volume={365},
  number={6459},
  pages={eaax1971},
  year={2019},
  publisher={American Association for the Advancement of Science},
  doi={10.1126/science.aax1971},
  note={9 pages}
}

@misc{monocle3trajectories,
  title={Constructing single-cell trajectories},
  author={Cole Trapnell}, 
  year={2022},
  howpublished={\url{https://cole-trapnell-lab.github.io/monocle3/docs/trajectories/}},
  note={Accessed: 2024-03-24}
}

@article{monocle3,
  title={The single-cell transcriptional landscape of mammalian organogenesis},
  author={Cao, Junyue and Spielmann, Malte and Qiu, Xiaojie and Huang, Xingfan and Ibrahim, Daniel M and others},
  journal={Nature},
  volume={566},
  number={7745},
  pages={496--502},
  year={2019},
  publisher={Nature Publishing Group UK London},
  doi={10.1038/s41586-019-0969-x}
}

@misc{mnist,
  author = {LeCun, Yann and Cortes, Corinna and J.C. Burges, Christopher},
  year = {1999},
  title = {The {MNIST} database of handwritten digits},
  howpublished = {\url{https://www.openml.org/search?type=data&id=554}},
  note = {Accessed: 2024-3-26}
}

@inproceedings{Lang95,
    author = {Ken Lang},
    title = {{NewsWeeder}: Learning to filter netnews},
    year = {1995},
    booktitle = {Proc. ICML},
    pages = {331--339},
  doi       = {10.1016/B978-1-55860-377-6.50048-7}
}

@article{blei2003latent,
  title={Latent {Dirichlet} allocation},
  author={Blei, David M and Ng, Andrew Y and Jordan, Michael I},
  journal={J. Mach. Learn. Res.},
  volume={3},
  number={Jan},
  pages={993--1022},
  year={2003},
  note={\url{https://proceedings.neurips.cc/paper_files/paper/2001/file/296472c9542ad4d4788d543508116cbc-Paper.pdf}}
}

@book{ronchetti2009robust,
  title={Robust Statistics},
  author={Huber, Peter J and Ronchetti, Elvezio M},
  year={2009},
  publisher={John Wiley \& Sons Hoboken},
  doi       = {10.1002/9780470434697}
}

@article{xu2010robust,
  title={Robust {PCA} via outlier pursuit},
  author={Xu, Huan and Caramanis, Constantine and Sanghavi, Sujay},
  journal={Adv. Neural Inf. Process. Syst.},
  volume={23},
  year={2010}
}

@article{McInnes2018, doi = {10.21105/joss.00861}, url = {https://doi.org/10.21105/joss.00861}, year = {2018}, publisher = {The Open Journal}, volume = {3}, number = {29}, pages = {861}, author = {McInnes, Leland and Healy, John and Saul, Nathaniel and Großberger, Lukas}, title = {{UMAP}: Uniform Manifold Approximation and Projection}, journal = {J. Open Source Softw.} }

@article{candes2011robust,
author = {Cand\`{e}s, Emmanuel J. and Li, Xiaodong and Ma, Yi and Wright, John},
title = {Robust principal component analysis?},
year = {2011},
issue_date = {May 2011},
publisher = {ACM},
volume = {58},
number = {3},
url = {https://doi.org/10.1145/1970392.1970395},
doi = {10.1145/1970392.1970395},
journal = {J. ACM},
month = jun,
articleno = {11},
numpages = {37}
}

@misc{robust-pca-github,
author={{Deep Ganguli}},
title={Robust {PCA}},
year={2019},
howpublished={\url{https://github.com/dganguli/robust-pca}},
note={Accessed: 2026-3-28}
}

@article{raschka2020machine,
  title={Machine learning in {Python}: Main developments and technology trends in data science, machine learning, and artificial intelligence},
  author={Raschka, Sebastian and Patterson, Joshua and Nolet, Corey},
  journal={Information},
  volume={11},
  number={4},
  pages={193},
  year={2020},
  publisher={mdpi},
  doi={https://doi.org/10.3390/info11040193}
}

@misc{tensorpqm,
author={Machado, Alister},
title={Accelerated Projection Quality Metrics},
year={2024},
howpublished={\url{https://github.com/amreis/tf-projection-qm}},
note={Accessed: 2026-03-31}
}

@inproceedings{rauber2016visualizing,
author = {Rauber, Paulo E. and Falc{\~a}o, Alexandre X. and Telea, Alexandru C.},
title = {Visualizing time-dependent data using dynamic {t-SNE}},
year = {2016},
publisher = {Eurographics},
booktitle = {Proc. EuroVis (Short Papers)},
pages = {73–77},
numpages = {5}
}

@article{wang2022joint,
  title={Joint {t-SNE} for Comparable Projections of Multiple High-Dimensional Datasets},
  author={Wang, Yinqiao and Chen, Lu and Jo, Jaemin and Wang, Yunhai},
  journal={IEEE Trans. Vis. Comput. Graph.},
  volume={28},
  number={1},
  pages={623--632},
  year={2022},
  publisher={IEEE},
  doi={10.1109/TVCG.2021.3114765}
}

@article{atzberger2025large,
  title={A Large-Scale Sensitivity Analysis on Latent Embeddings and Dimensionality Reductions for Text Spatializations},
  author={Atzberger, Daniel and Cech, Tim and Scheibel, Willy and D{\"o}llner, J{\"u}rgen and Behrisch, Michael and Schreck, Tobias},
  journal={IEEE Trans. Vis. Comput. Graph.},
  volume={31},
  number={1},
  pages={305--315},
  year={2025},
  publisher={IEEE},
  doi={10.1109/TVCG.2024.3456308}
}

@inproceedings{bredius2022visual,
  title={Visual Exploration of Neural Network Projection Stability},
  author={Bredius, C. and Tian, Z. and Telea, A.},
  booktitle={Proc. MLVIS},
  year={2022},
  publisher={Eurographics},
  doi={https://doi.org/10.2312/mlvis.20221068},
  note={5 pages}
}

@article{vernier2020quantitative,
  title={Quantitative Evaluation of Time-Dependent Multidimensional Projection Techniques},
  author={Vernier, E. F. and Garcia, R. and da Silva, I. P. and Comba, J. L. D. and Telea, A. C.},
  journal={Comput. Graph. Forum},
  volume = {39},
  number = {3},
  pages = {241-252},
  year={2020},
  doi={10.1111/cgf.13977}
}

@book{gower2004procrustes,
  title={Procrustes Problems},
  author={Gower, John C and Dijksterhuis, Garmt B},
  volume={30},
  year={2004},
  publisher={Oxford University Press on Demand}
}

@article{faust2018dimreader,
  title={{DimReader}: Axis lines that explain non-linear projections},
  author={Faust, Rebecca and Glickenstein, David and Scheidegger, Carlos},
  journal={IEEE Trans. Vis. Comput. Graph.},
  volume={25},
  number={1},
  pages={481--490},
  year={2019},
  publisher={IEEE}
}

@article{fujiwara2019supporting,
  title={Supporting analysis of dimensionality reduction results with contrastive learning},
  author={Fujiwara, Takanori and Kwon, Oh-Hyun and Ma, Kwan-Liu},
  journal={IEEE Trans. Vis. Comput. Graph.},
  volume={26},
  number={1},
  pages={45--55},
  year={2020},
  publisher={IEEE}
}

@article{smelser2026scale,
  title={How scale breaks ``Normalized Stress''' and {KL} divergence: Rethinking quality metrics},
  author={Smelser, Kiran and Gunaratne, Kaviru and Miller, Jacob and Kobourov, Stephen},
  journal={IEEE Trans. Vis. Comput. Graph.},
  year={2026},
  volume={32},
  number={7},
  pages={5020-5033},
  doi={10.1109/TVCG.2026.3657654}
}

@article{fujiwara2025adversarial,
  title={Adversarial attacks on machine learning-aided visualizations},
  author={Fujiwara, Takanori and Kucher, Kostiantyn and Wang, Junpeng and Martins, Rafael M and Kerren, Andreas and Ynnerman, Anders},
  journal={J. Visualization},
  volume={28},
  number={1},
  pages={133--151},
  year={2025},
  publisher={Springer},
  doi={https://doi.org/10.1007/s12650-024-01029-2}
}

@article{fujiwara2019incremental,
  title={An incremental dimensionality reduction method for visualizing streaming multidimensional data},
  author={Fujiwara, Takanori and Chou, Jia-Kai and Shilpika, Shilpika and Xu, Panpan and Ren, Liu and Ma, Kwan-Liu},
  journal={IEEE Trans. Vis. Comput. Graph.},
  volume={26},
  number={1},
  pages={418--428},
  year={2020},
  doi={10.1109/TVCG.2019.2934433}
}

@article{sainburg2021parametric,
  title={Parametric {UMAP} embeddings for representation and semisupervised learning},
  author={Sainburg, Tim and McInnes, Leland and Gentner, Timothy Q},
  journal={Neural Computation},
  volume={33},
  number={11},
  pages={2881--2907},
  year={2021},
  doi={https://doi.org/10.1162/neco_a_01434}
}

\end{document}